\documentclass[11pt]{article}

\usepackage{shortcut}
\usepackage[round]{natbib}
\usepackage{times}
\usepackage{dsfont}
\usepackage{parskip}
\usepackage{placeins}
\usepackage[margin=1in]{geometry}

\graphicspath{{figures/}{./}}

\newcommand{\ttran}[1]{#1^{\Lsh}}
\newcommand{\Cm}{\mathcal C_m}
\newcommand{\Sw}{\mathcal S_w}
\newcommand{\Cmw}{\mathcal C_m^{(w)}}
\newcommand{\mDICOD}{DiCoDiLe-$Z$}
\newcommand {\bO} [1] {\mathcal O(#1)}

\binoppenalty=\maxdimen
\relpenalty=\maxdimen

\title{Distributed Convolutional Dictionary Learning (DiCoDiLe): \\ Pattern Discovery in Large Images and Signals}
\author{
       Moreau Thomas$^1$\footnote{
           Corresponding author \href{mailto:thomas.moreau@inria.fr}{thomas.moreau@inria.fr}},
       Alexandre Gramfort$^1$\\[.3em]
       $^1$INRIA, Université Paris Saclay, Saclay, France}

\begin{document}

\maketitle

\begin{abstract}

    Convolutional dictionary learning (CDL) estimates shift invariant basis adapted to multidimensional data. CDL has proven useful for image denoising or inpainting, as well as for pattern discovery on multivariate signals. As estimated patterns can be positioned anywhere in signals or images, optimization techniques face the difficulty of working in extremely high dimensions with millions of pixels or time samples, contrarily to standard patch-based dictionary learning. To address this optimization problem, this work proposes a distributed and asynchronous algorithm, employing locally greedy coordinate descent and an asynchronous locking mechanism that does not require a central server. This algorithm can be used to distribute the computation on a number of workers which scales linearly with the encoded signal's size. Experiments confirm the scaling properties which allows us to learn patterns on large scales images from the Hubble Space Telescope.

\end{abstract}

\section{Introduction}
\label{sec:dicod:context}

The sparse convolutional linear model has been used successfully for various signal and image processing applications. It was first developed by~\citet{Grosse2007} for music recognition and has then been used extensively for denoising and inpainting of images \citep{Kavukcuoglu2013,Bristow2013,Wohlberg2014}. The benefit of this model over traditional sparse coding technique is that it is  shift-invariant. Instead of local patches, it considers the full signals which allows for a sparser reconstruction, as not all shifted versions of the same patches are present in the estimated dictionary~\citep{Bergstra-etal:11}.

Beyond denoising or inpainting, recent applications used this shift-invariant model as a way to learn and localize important patterns in signals or images. \citet{Yellin2017} used convolutional sparse coding to count the number of blood-cells present in holographic lens-free images. \citet{Gramfort2017} and \citet{Dupre2018} learned recurring patterns in univariate and multivariate signals from neuroscience. This model has also been used with astronomical data as a way to denoise \citep{Starck2005} and localize predefined patterns from space satellite images \citep{DelAguilaPla2018a}.

Convolutional dictionary learning estimates the parameters of this model. It boils down to an optimization problem, where one needs to estimate the patterns (or atoms) of the basis as well as their activations, \ie{} where the patterns are present in the data. The latter step, which is commonly referred to as convolutional sparse coding (CSC), is the critical bottleneck when it comes to applying CDL to large data. \citet{Chalasani2013} used an accelerated proximal gradient method based on Fast Iterative Soft-Thresholding Algorithm (FISTA;~\citealt{Beck2009} that was adapted for convolutional problems. The Fast Convolutional Sparse Coding of \citet{Bristow2013} is based on Alternating Direction Method of Multipliers (ADMM). Both algorithms compute a convolution on the full data at each iteration. While it is possible to leverage the Fast Fourier Transform to reduce the complexity of this step \citep{Wohlberg2016}, the cost can be prohibitive for large signals or images. To avoid such costly global operations, \citet{Moreau2018} proposed to use locally greedy coordinate descent (LGCD) which requires only local computations to update the activations. Based on Greedy Coordinate Descent (GCD), this method has a lower per-iteration complexity and it relies on local computation in the data, like a patch based technique would do, yet solving the non-local problem.

To further reduce the computational burden for CDL, recent studies consider different parallelization strategies. \citet{Skau2018} proposed a method based on consensus ADMM to parallelize the CDL optimization by splitting the computation across the different atoms. This limits the number of cores that can be used to the number of atoms in the dictionary, and does not significantly reduce the burden for very large signals or images. An alternative parallel algorithm called DICOD, proposed by \citet{Moreau2018} distributes GCD updates when working with 1d signals. Their results could be used with multidimensional data such as images, yet only by splitting the data along one dimension. This limits the gain of their distributed optimization strategy when working with images. Also, the choice of GCD for the local updates make the algorithm inefficient when used with low number of workers.

In the present paper, we propose a distributed solver for CDL, which can fully leverage multidimensional data. For the CSC step, we present a coordinate selection rule based on LGCD that works in the asynchronous setting without a central server. We detail a soft-lock mechanism ensuring the convergence with a grid of workers, while splitting the data in all convolutional dimensions. Then, we explain how to also leverage the distributed set of workers to further reduce the cost of the dictionary update, by precomputing in parallel some quantities used in the gradient operator. Extensive numerical experiments confirm the performance of our algorithm, which is then used to learn patterns from a large astronomical image.

In \autoref{sec:cdl}, we describe the convolutional linear model and its parameter estimation through CDL. Then, \autoref{sec:lgcd} details the LGCD algorithm for convolutional sparse coding (CSC). Our distributed algorithm DiCoDiLe is introduced in \autoref{sec:dicodil}. \autoref{sec:expes} presents results on simulations and images.

\section{Convolutional Dictionary Learning}
\label{sec:cdl}

\paragraph{Notations.}
For a vector $U\in\Rset^P$ we denote $U_p\in\Rset$ its $p^{th}$ coordinate. For finite domain $\Omega =\prod_{i=1}^d [0, T_i[$, where $[0, T_i[$ are the integers from $0$ to $T_i-1$, $|\Omega|$ is the size of this domain $\prod_{i=1}^d T_i$. We denote $\Xset_\Omega^P$ (resp. $\Xset_\Omega^{P\times Q}$) the space of observations defined on $\Omega$ with values in $\Rset^P$ (resp. $\Rset^{P\times Q}$). For instance, $\Xset_\Omega^3$ with $d=2$ is the space of RGB-images with height and width $T_1, T_2$. The value of an observation $X \in \Xset_\Omega^P$ at position $\omega\in\Omega$ is given by $X[\omega] = X[\omega_1, \dots, \omega_d] \in \Rset^P$, and its restriction to the $p^{th}$ coordinate at each position is denoted $X_p \in \Xset_\Omega^1$. All signals are 0-padded, \ie{} for $\omega\not\in\Omega$, $X[\omega] = 0$. The convolution operator is denoted $*$ and is defined for $X \in \Xset_\Omega^K$ and $\pmb  Y \in \Xset_\Omega^{K\times P}$ as the signal $X*\pmb Y \in \Xset_\Omega^P$ with
\begin{equation}
(X * \pmb Y)[\omega] = \sum_{\tau \in \Omega} X[\omega-\tau]\cdot \pmb Y[\tau] ~,~~\forall \omega \in \Omega \enspace .
\end{equation}
For any signal $X \in \Xset_\Omega^P$, the reversed signal is defined as $\ttran{X}[\omega] = X[T_1-\omega_1, \dots, T_d- \omega_d]\tran{}$ and the $p$-norm is defined as $\|X\|_p = \left(\sum_{\omega \in \Omega}\|X[\omega]\|_p^p\right)^{1/p}$. We will also denote ST the soft-thresholding operator defined as
\[
    \text{ST}(u, \lambda) = \text{sign}(u)\max(|u| - \lambda, 0)~.
\]

\paragraph{Convolutional Linear Model.}

\begin{figure}[t!]
    \begin{center}
        \centerline{\includegraphics[width=.8\textwidth]{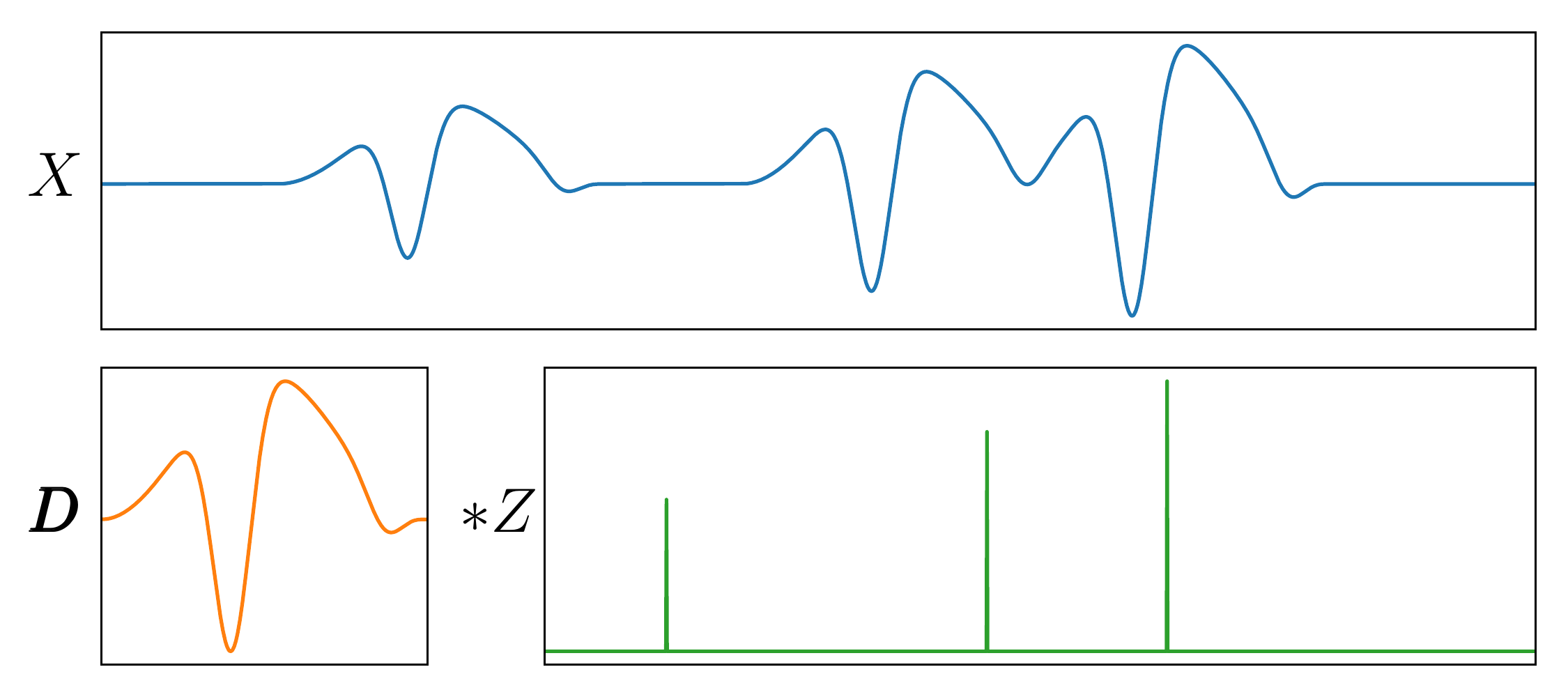}}
        \caption{Decomposition of a noiseless univariate signal $X$ (\emph{blue}) as the convolution $Z*\pmb D$ between a temporal pattern $\pmb D$ (\emph{orange}) and a sparse activation signal $Z$ (\emph{green}).}
        \label{fig:cdl:presentation}
    \end{center}
\end{figure}

For an observation $X\in\Xset_\Omega^P$, the convolutional linear model reads
\begin{equation}\label{eq:model}
X =  Z * \pmb D + \xi~,
\end{equation}
with $\pmb D \in \Xset_\Theta^{K\times P}$ a dictionary of $K$ atoms with the same number of dimensions $P$ on a domain $\Theta = \prod_{i=1}^d[0, L_i[ \subset \Omega$ and  $Z \in \Xset_\Omega^K$ the set of $K$ activation vectors associated with this dictionary. $\xi \in \Xset_\Omega^P$ is an additive noise term which is typically considered Gaussian and white. The dictionary elements $\pmb D_k$ are patterns present in the data and typically have a much smaller support $\Theta$ than the full domain $\Omega$. The activations $Z_k$ encode the localization of the pattern $k$ in the data. As patterns typically occur at few locations in each observation, the coding signal $Z$ is considered to be sparse. \autoref{fig:cdl:presentation} illustrates the decomposition of a temporal signal with one pattern.

The parameters of this model are estimated by solving the following optimization problem, called convolutional dictionary learning (CDL)
\begin{equation}
    \label{eq:cdl}
    \min_{\substack{Z, \pmb D\\ \|\pmb D_k\| \le 1}}
        \underbrace{\frac{1}{2}\left\| X - Z * \pmb D \right\|_2^2}_{F(Z, \pmb D)}
            + \underbrace{\lambda\|Z\|_1}_{G(Z)}~.
\end{equation}
where the first term measures how well the model fits the data, the second term enforces a sparsity constraint on the activation, and
$\lambda > 0$ is a regularization parameter that balances the two. The constraint $\|D_k\|_2 \le 1$ is used to avoid the scale ambiguity in our model -- due to the $\ell_1$ constraint. Indeed, the $\ell_1$-norm of $Z$ can be made arbitrarily small by rescaling $D$ by $\alpha \in \Rset_+^*$ and $Z$ by $\frac{1}{\alpha}$. The minimization \autoref{eq:cdl} is not jointly convex in $(Z, \pmb D)$ but it is convex in each of its coordinates. It is usually solved with an alternated minimization algorithm, updating at each iteration one of the block of coordinate, $Z$ or $\pmb D$.

\paragraph{Convolutional Sparse Coding (CSC).}
Given a dictionary of patterns $\pmb D$, CSC aims to retrieve the sparse decomposition $Z^*$ associated to the signal $X$. When minimizing only over $Z$, \autoref{eq:cdl} becomes
\begin{equation}
\label{eq:csc}
    Z^\star = \argmin_{Z} \frac{1}{2} \left\|X - Z* \pmb D\right\|_2^2
        + \lambda\left\|Z\right\|_1 \enspace .
\end{equation}
The problem in \autoref{eq:csc} boils down to a LASSO problem with a Toeplitz design matrix $\pmb D$. Therefore, classical LASSO optimization techniques can easily be applied with the same convergence guarantees. \citet{Chalasani2013} adapted FISTA \citep{Beck2009} exploiting the convolution in the gradient computation to save time. Based on the ADMM algorithm \citep{Gabay1976}, \citet{Bristow2013} solved the CSC problem with a fast computation of the convolution in the Fourier domain. For more details on these techniques, see \citet{Wohlberg2016} and references therein. Finally, \citet{Kavukcuoglu2013} proposed an efficient adaptation of the greedy coordinate descent \citep{Osher2009} to efficiently solve the CSC, which has been later refined by \citet{Moreau2018} with locally greedy coordinate selection (LGCD).

It is useful to note that for $\lambda \ge \|X*\ttran{\pmb D}\|_\infty$, $0$ minimizes the equation \autoref{eq:csc}. The first order condition for \autoref{eq:csc} reads
\begin{equation}
    \nabla_Z F(0, \pmb D) = X * \ttran{\pmb D} \in \partial G(0) = [-\lambda, \lambda]^{K|\Omega|}~.
\end{equation}
This condition is always verified if $\lambda > \lambda_{\max} = \|X*\ttran{\pmb D}\|_\infty$. In the following, the regularization parameter $\lambda$ is set as a fraction of this value.

\paragraph{Dictionary Update.}
\label{par:contrib:dict_update}

Given a activation signal $Z^{(q)}$, the dictionary update aims at improving how the model reconstructs the signal $X$ using the fixed codes, by solving
\begin{equation}
    \label{eq:dict_update}
    \pmb D^* = \argmin_{\substack{\pmb D\\\|\pmb D_k\|_2 \le 1}} \frac{1}{2} \left\|X - Z* \pmb D\right\|_2^2 \enspace .
\end{equation}
This problem is smooth and convex and can be solved using classical algorithms. The block coordinate descent for dictionary learning \citep{Mairal2010} can be easily extended to the convolutional cases. Recently, \citet{Yellin2017} proposed to adapt the K-SVD dictionary update method \citep{Aharon2006} to the convolutional setting. Classical convex optimization algorithms, such as projected gradient descent (PGD) and its accelerated version (APGD), can also be used for this step \citep{Combettes2011}. These last two algorithms are often referred to as ISTA and FISTA \citep{Chalasani2013}.

\section{Locally Greedy Coordinate Descent for CSC}
\label{sec:lgcd}

\begin{algorithm}[t]
    \begin{algorithmic}[1]
        \STATE \textbf{Input: }$\pmb D, X$, parameter $\epsilon >  0$, sub-domains $\Cm$,
        \STATE Initialization: $\forall (k,\omega) \in [1, K]\times\Omega,~~ $  \\
        $Z_k[\omega] = 0,~~\beta_k[\omega] = \left(\ttran{\pmb D_k} * X\right)[\omega]$
        \REPEAT
        \FOR{$m = 1\dots M$}
        \STATE $\forall(k, \omega) \in [1,K]\times\Cm{}$,\hskip1em
        $\displaystyle Z'_k[\omega] = \frac{1}{\|\pmb D_k\|_2^2}\text{ST}(\beta^{(q)}_k[\omega], \lambda)~,$
        \STATE Choose $\displaystyle(k_0, \omega_0) = \arg\max_{(k, \omega)\in[1, K]\times\Cm{}} |\Delta Z_k[\omega]|$
        \STATE Update $\beta$ using \autoref{eq:beta_update} and $Z_{k_0}[\omega_0] \leftarrow Z'_{k_0}[\omega_0]$
        \ENDFOR
        \UNTIL{$\|\Delta Z\|_\infty < \epsilon$}

    \end{algorithmic}
    \caption{Locally Greedy Coordinate Descent}
    \label{alg:lgcd}
\end{algorithm}

Coordinate descent (CD) is an algorithm which updates a single coefficient at each iteration. For \autoref{eq:csc}, it is possible to compute efficiently, and in closed form, the optimal update of a given coordinate when all the others are kept fixed. Denoting $q$ the iteration number, and $Z'_{k_0}[\omega_0]$ (atom $k_0$ at \mbox{position $\omega_0$}) the updated version of $Z^{(q)}_{k_0}[\omega_0]$, the update reads
\begin{equation} \label{eq:pb_coord}
Z'_{k_0}[\omega_0]
= \frac{1}{\|\pmb D_{k_0}\|_2^2}\text{ST}(\beta^{(q)}_{k_0}[\omega_0], \lambda) \enspace ,
\end{equation}
where $\beta^{(q)}$ is an auxiliary variable in $\Xset^K_\Omega$ defined for $(k, \omega)\in[1, K]\times\Omega$ as
\[
    \smeq\displaystyle \beta_{k}^{(q)}[\omega] =
        \left(\left(X - Z^{(q)}*\pmb D + Z_{k}^{(q)}[\omega] e_\omega*\pmb D_{k}\right)
              *\ttran{\pmb D_{k}}\right)[\omega]~,
\]
and where $e_\omega$ is a dirac (canonical vector) with value 1 in $\omega$ and 0 elsewhere. To make this update
efficient it was noticed by \citet{Kavukcuoglu2013} that if the coordinate $Z_{k_0}[\omega_0]$ is updated with an additive update $\Delta Z_{k_0}[\omega_0] = Z'_{k_0}[\omega_0] - Z^{(q)}_{k_0}[\omega_0]$, then it is possible to cheaply obtain $\beta^{(q+1)}$ from $\beta^{(q)}$ using the relation
\begin{equation}\label{eq:beta_update}
    \beta_k^{(q+1)}[\omega] = \beta_k^{(q)}[\omega] -
        (\pmb D_{k_0} * \ttran{\pmb D_k})[\omega-\omega_0]
            \Delta Z^{(q)}_{k_0}[\omega_0]~,
\end{equation}
for all $(k, \omega) \neq (k_0, \omega_0)$. As the support of $(\pmb D_{k_0} * \ttran{\pmb D_k})[\omega]$ is $\prod_{i=1}^d [-L_i + 1, L_i[$, equation \autoref{eq:beta_update} implies that, after an update in $\omega_0$, the value $\beta^{(q+1)}_k$ only changes in the neighborhood $\mathcal V(\omega_0)$ of $\omega_0$, where
\begin{equation}
     \mathcal V(\omega_0) = \prod_{i=1}^d
            [(\omega_0)_i - L_i + 1, (\omega_0)_i + L_i[ \enspace .
\end{equation}
We recall that $L_i$ is the size of an atom in the direction $i$. This means that only $\mathcal O(2^dK|\Theta|)$ operations are needed to maintain $\beta$ up-to-date after each $Z$ update. The procedure is run until $\max_{k, t}|\Delta Z_{k}[t]|$ becomes smaller than a specified tolerance parameter $\epsilon \ge 0$.

The selection of the updated coordinate $(k_0, \omega_0)$ can follow different strategies. Cyclic updates \citep{Friedman2007} and random updates \citep{Shalev2009} are efficient strategies which only require to access one value of \autoref{eq:pb_coord}. They have a $\bO{1}$ computational complexity. \citet{Osher2009} propose to select greedily the coordinate which is the farther from its optimal value. In this case, the coordinate is chosen as the one with the largest additive update  $\max_{(k, \omega)}|\Delta Z_{k}[\omega]|$. The quantity $\Delta Z_k[\omega]$ acts as a proxy for the cost reduction obtained with this update.

This strategy is computationally more expensive, with a cost of $\bO{K|\Omega|}$. Yet, it has a better convergence rate \citep{Nutini, Karimireddy2018} as it updates in priority the important coordinates. To reduce the iteration complexity of greedy coordinate selection while still selecting more relevant coordinates, \citet{Moreau2018} proposed to select the coordinate in a locally greedy fashion. The domain $\Omega$ is partitioned in $M$ disjoint sub-domains $\Cm$: $\Omega = \cup_m \Cm$ and $\Cm \cap \mathcal C_{m'} = \emptyset$ for $m \neq m'$. Then, at each iteration $q$, the coordinate to update is selected greedily on the $m$-th sub-domain $\Cm$
\[
    (k_0, \omega_0) = \argmax_{(k, \omega)\in[1, K]\times\mathcal C_m} |\Delta Z_k[\omega]| \enspace ,
\]
\mbox{with $m = q \mod M$} (cyclic rule).

This selection differs from the greedy selection as it is only locally selecting the best coordinate to update. The iteration complexity of this strategy is therefore linear with the size of the sub-segments $\bO{K|\mathcal C_m|}$. If the size of the sub-segments is $1$, this algorithm is equivalent to a cyclic coordinate descent, while with $M=1$ and $\mathcal C_1 = \Omega$ it boils down to greedy updates. The selection of the sub-segment size is therefore a trade-off between the cyclic and greedy coordinate selection. By using sub-segments of size $2^d|\Theta|$, the computational complexity of selecting the coordinate to update and the complexity of the update of $\beta$ are matched. The global iteration complexity of LGCD becomes $\bO{K|\Theta|}$. \autoref{alg:lgcd} summarizes the computations for LGCD.

\section{Distributed Convolutional Dictionary Learning (DiCoDiLe)}
\label{sec:dicodil}

 \begin{algorithm}[t]
    \begin{algorithmic}[1]
        \STATE \textbf{Input: }$\pmb D^{(0)}, X$, stopping criterion $\nu$.
        \REPEAT
        \STATE Compute $Z^{(q+1)}$ with \mDICOD{}$(X, \pmb D^{(q)}, W)$.
        \STATE Compute $\phi$ and $\psi$ with \autoref{eq:phi} and $W$ workers.
        \STATE Update $\pmb D^{(q+1)}$ with PGD and armijo backtracking line-search.
        \UNTIL{Cost variation is smaller than $\nu$ }
    \end{algorithmic}
    \caption{DiCoDiLe with $W$ workers}
    \label{alg:dicodile}
\end{algorithm}

In this section, we introduce DiCoDiLe, a distributed algorithm for convolutional sparse coding. Its step are summarized in \autoref{alg:dicodile}. The CSC part of this algorithm will be referred to as \mDICOD{} and extend the DICOD algorithm to handle multidimensional data such as images, which were not covered by the original algorithm. The key to ensure the convergence with multidimensional data is the use of asynchronous soft-locks between neighboring workers.  We also propose to employ a locally greedy CD strategy which boosts running time, as demonstrated in the experiments. The dictionary update relies on smart pre-computation which make the computation of the gradient of \autoref{eq:dict_update} independent of the size of $\Omega$.

\subsection{Distributed sparse coding with LGCD}
\label{sub:dicodil:dicod}

 \begin{algorithm}[t]
    \begin{algorithmic}[1]
        \STATE \textbf{Input: }$\pmb D, X$, parameter $\epsilon >  0$,  sub-domains $\Sw$,
        \STATE \textbf{In parallel } for $w=1\cdots W$
        \STATE Compute a partition $\{\Cmw{}\}_{m=1}^M$ of the worker domain $\Sw$ with sub-domains of size $2^d|\Theta|$.
        \STATE Initialize $\beta_k[t]$ and $Z_k[t]$ $~\forall(k,t) \in [1, K]\times\Sw{}$,
        \REPEAT
        \FOR {m=1\dots M}
        \STATE Receive messages and update $Z$ and $\beta$ with \autoref{eq:beta_update}
        \STATE Choose $\displaystyle(k_0, \omega_0) = \argmax_{(k, \omega)\in[1, K]\times\Cmw{}} |\Delta Z_k[\omega]|$
        \IF{$\displaystyle|\Delta Z_{k_0}[\omega_0]| < \max_{(k, \omega)\in[1, K]\times\mathcal V(\omega_0)} |\Delta Z_k[\omega]$}
        \STATE The coordinate is soft-locked, \textbf{goto 6}
        \ENDIF
        \STATE Update $\beta$ with \autoref{eq:beta_update} and $Z_{k_0}[\omega_0]\leftarrow{}Z'_{k_0}[\omega_0]$
        \IF{ $\omega_0 \in \mathcal B_{2L}(\Sw)$ }
        \STATE ~~~~Send $(k_0, \omega_0, \Delta Z_{k_0}[\omega_0])$ to neighbors
        \ENDIF
        \ENDFOR
        \UNTIL{global convergence $\|\Delta Z\|_\infty < \epsilon$ }
    \end{algorithmic}
    \caption{\mDICOD{} with $W$ workers}
    \label{alg:dicod}
\end{algorithm}

For convolutional sparse coding, the coordinates that are far enough -- compared to the size of the dictionary -- are only weakly dependent. It is thus natural to parallelize CSC resolution by splitting the data in continuous sub-domains. It is the idea of the DICOD algorithm proposed by \citet{Moreau2018} for one-dimensional signals.

Given $W$ workers, DICOD partitions the domain $\Omega$ in $W$ disjoint contiguous time intervals $\mathcal S_w$.

Then each worker $w$ runs asynchronously a greedy coordinate descent algorithm on $\Sw$. To ensure the convergence, \citet{Moreau2018} show that it is sufficient to notify the neighboring workers if a local update changes the value of $\beta$ outside of $\Sw$.

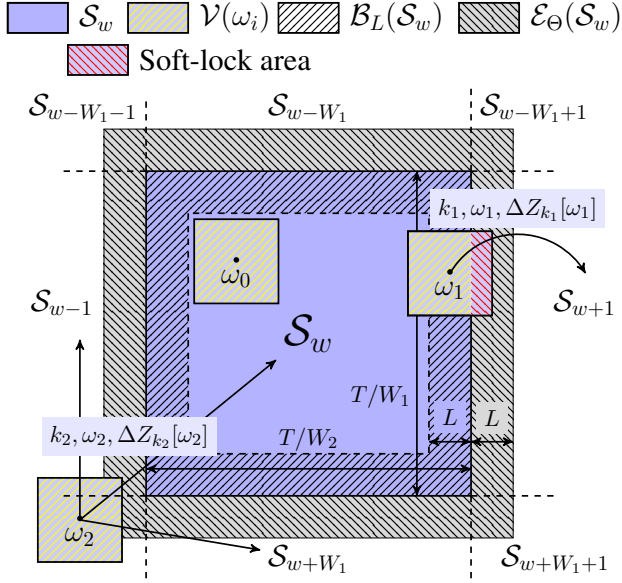
\begin{figure}[tp]
        \begin{minipage}{\columnwidth}
        \begin{minipage}{0.48\columnwidth}
    \scalebox{.8}{
    \begin{tikzpicture}

    \usetikzlibrary{patterns}

    \tikzset{
        >=stealth',
        bL/.style={
            postaction={
                pattern=north east lines,
                pattern color=black
            }
        },
        eL/.style={
            fill=gray!30,
            postaction={
                pattern=north west lines,
                pattern color=black, dashed
            }
        },
        vL/.style={
            fill=blue!20,
            postaction={
                pattern=north east lines,
                pattern color=yellow
            }
        },
        sL/.style={
            fill=blue!20,
            postaction={
                pattern=north west lines,
                pattern color=red
            }
        },
        Sw/.style={
            fill=blue!30,
            thick
        }
    };

    \pgfmathsetmacro{\T}{2}
    \pgfmathsetmacro{\L}{.7}
    \pgfmathsetmacro{\lmiddle}{3.25}

    \draw [eL] (-\T-2*\L, \T+2*\L) -- (-\T-2*\L, -\T-2*\L) -- (\T+2*\L, -\T-2*\L) -- (\T+2*\L, \T+2*\L) -- cycle;
    \draw [Sw, bL] (-\T-\L, \T+\L) -- (-\T-\L, -\T-\L) -- (\T+\L, -\T-\L) -- (\T+\L, \T+\L) -- cycle;
    \draw [Sw, dashed] (-\T, \T) -- (-\T, -\T) -- (\T, -\T) -- (\T, \T) -- cycle;

    \node (Sw) {\huge$\mathcal S_{w}$};
    \draw (-\T-3*\L, .5) node {\Large$\mathcal S_{w-1}$};
    \draw (\T+3.7*\L, .5) node {\Large$\mathcal S_{w+1}$};
    \draw (-\T-2.5*\L, \T+2.5*\L) node {\Large$\mathcal S_{w - W_1 - 1}$};
    \draw (0, \T+2.5*\L) node {\Large$\mathcal S_{w - W_1}$};
    \draw (\T+2.5*\L, \T+2.5*\L) node {\Large$\mathcal S_{w - W_1 + 1}$};
    \draw (0, -\T-2.5*\L) node {\Large$\mathcal S_{w + W_1}$};
    \draw (\T+3*\L, -\T-2.5*\L) node {\Large$\mathcal S_{w + W_1 +1}$};

    \draw [<->, thick] (-\T-\L, -2.25) -- node [above] (anchor) {} (\T+\L, -2.25);
    \node [above=-.3em of anchor] {$T/W_2$};
    \draw [<->, thick] (1.8, -\T-\L) -- node [left] (anchor) {} (1.8, \T+\L);
    \node [below=2em of anchor] (anchor) {};
    \node [left=-.8em of anchor] {$T/W_1$};
    \draw [<->, thick] (\T+\L, -1.8) -- node [above] (anchor) {} (\T, -1.8);
    \node [above=-.4em of anchor, Sw] {$L$};
    \draw [<->, thick] (\T+\L, -1.8) -- node [above] (anchor) {} (\T+2*\L, -1.8);
    \node [above=-.4em of anchor, fill=gray!30] {$L$};

    \foreach \x/\y/\i in {-1.2/1.2/0, \T+.5*\L/1/1, -\T-2*\L-.4/-\T-\L-.4/2}{
        \draw [vL] (-\L+\x, -\L+\y) rectangle (\L+\x, \L+\y);
        \draw [thick] (-\L+\x, -\L+\y) rectangle (\L+\x, \L+\y);
        \draw (\x, \y) node [below] {\Large$\omega_\i$} node {\Huge$\cdot$};

        "\ifthenelse{\i>0}{
            \ifthenelse{\i=1}{

                \draw [blue!10, sL] (\x + \L, \y + \L) rectangle (\x+.5*\L, \y-\L);
                \draw [thick] (\x - \L, \y - \L) rectangle (\x + \L, \y + \L);

                \draw [->, thick] (\x, \y) arc (150:30:1.3) node [midway, above] (w1) {};
                \node[above=-.5em of w1, fill=blue!10] {$k_\i, \omega_\i, \Delta Z_{k_\i}[\omega_\i]$};
            }{

                \draw [->, thick] (\x, \y) -- (Sw);
                \draw [->, thick] (\x, \y) -- (\x+3, \y-.5);

                \draw [->, thick] (\x, \y) -- (\x, \y+3) node [midway] (w1) {};
                \node[above=-1em of w1] (anchor) {};
                \node [right=-2em of anchor, fill=blue!10] {$k_\i, \omega_\i, \Delta Z_{k_\i}[\omega_\i]$};
            }
        }{}"
    }

    \draw [dashed, thick] (-\T - \L, \T + \L) -- (-\T - \L, \T+2.8*\L);
    \draw [dashed, thick] (-\T - \L, \T + \L) -- (-\T - 3*\L, \T + \L);
    \draw [dashed, thick] (\T + \L, \T + \L) -- (\T + \L, \T + 3*\L);
    \draw [dashed, thick] (\T + \L, \T + \L) -- (\T + 3*\L, \T + \L);
    \draw [dashed, thick] (\T + \L, -\T - \L) -- (\T + \L, -\T - 3*\L);
    \draw [dashed, thick] (\T + \L, -\T - \L) -- (\T+3*\L, -\T - \L);
    \draw [dashed, thick] (-\T - \L, -\T - \L) -- (-\T - \L, -\T - 3*\L);
    \draw [dashed, thick] (-\T - \L, -\T - \L) -- (-\T - 3*\L, -\T - \L);

    \pgfmathsetmacro{\htext}{\T+\L + 2.55}
    \pgfmathsetmacro{\hstart}{\T+\L + 2.3}
    \pgfmathsetmacro{\hend}{\T+\L + 2.8}
    \draw [Sw] (-5, \hstart) rectangle  (-4, \hend);
    \draw [black, thick] (-5, \hstart) rectangle  (-4, \hend);
    \draw (-3.95, \htext) node [right] {\Large$\mathcal S_w$};
    \draw [vL] (-3, \hstart) rectangle  (-2, \hend);
    \draw [black, thick] (-3, \hstart) rectangle  (-2, \hend);
    \draw (-1.95,  \htext) node [right] {\Large$\mathcal V(\omega_i)$};
    \draw [bL] (-0.5, \hstart) rectangle  (0.5, \hend);
    \draw [black, thick] (-0.5, \hstart) rectangle  (0.5, \hend);
    \draw (0.55, \htext) node [right] {\Large$\mathcal B_L(\mathcal S_w)$};
    \draw [eL] (2.5, \hstart) rectangle  (3.5, \hend);
    \draw [black, thick] (2.5, \hstart) rectangle  (3.5, \hend);
    \draw (3.55, \htext) node [right] {\Large$\mathcal E_\Theta(\mathcal S_w)$};
    \draw [sL] (-4, \hstart - .7) rectangle  (-3, \hend - .7);
    \draw [black, thick] (-4, \hstart - .7) rectangle  (-3, \hend - .7);
    \draw (-2.95, \htext - .7) node [right] {\Large Soft-lock area};
    \end{tikzpicture}
}
\end{minipage}
\hfill
\begin{minipage}{0.49\columnwidth}
\vspace{-0.18in}

\caption{
    Communication process in \mDICOD{} with $d=2$ and 9 workers centered around worker $w$. The update in $\omega_0$ is independent of the other workers. The update in $\omega_1$ is performed if no better coordinate update is possible in the soft-lock area $\mathcal V(\omega_1) \cap \mathcal S_{w+1}$ (\emph{red hatched area}). If accepted, the worker $w+1$ needs to be notified. The update in $\omega_2$ changes the value of the optimal updates in the $\Theta$-extension of the other workers sub-domains. Thus it needs to notify all the neighboring workers.
}
\label{fig:communication}
\end{minipage}
\end{minipage}
\end{figure}

Let us consider a sub-domain, $\Sw = \prod_{i=1}^d [l_i, u_i[$. The $\Theta$-border is
\begin{equation}
    \mathcal B_L(\Sw) = \prod_{i=1}^d [l_i, l_i + L_i[\cup [u_i - L_i, u_i[~.
\end{equation}
In case that a $Z$ update affects $\beta$ in the $\Theta$-border $\mathcal B_L(\Sw)$ of $\Sw$, then one needs to notify other workers $w'$ whose domain overlap with the update of $\beta$ \ie $\mathcal V(\omega_0) \cap \mathcal S_{w'} \neq \emptyset$. It is done by sending the triplet $(k_0, \omega_0, \Delta Z_{k_0}[\omega_0])$ to these workers, which can then update $\beta$ using formula \autoref{eq:beta_update}. \autoref{fig:communication} illustrates this communication process on a 2D example. There is few inter-processes communications in this distributed algorithm as it does not rely on centralized communication, and a worker only communicates with its neighbors. Moreover, as the communication only occurs when $\omega_0 \in \mathcal{B}_L(\Sw)$, a small number of iterations need to send messages if $|\mathcal{B}_L(\Sw)| \ll |\Sw|$.

As a worker can be affected by its neighboring workers, the stopping criterion of CD cannot be applied independently in each worker. To reach a consensus, the convergence is considered to be reached once no worker can offer a change on a $Z_k[\omega]$ that is higher than $\epsilon$. Workers that reach this state locally are paused, waiting for incoming communication or for the global convergence to be reached.

\paragraph{\mDICOD} This algorithm, described in \autoref{alg:dicod}, is distributed over  $W$ workers which update asynchronously the coordinates of $Z$. The domain $\Omega$ is once also partitioned with sub-domain $\Sw$ but unlike DICOD, the partitioning is not restricted to be chosen along one direction. Each worker $w \in [1, W]$ is in charge of updating the coordinates of one sub-domain $\Sw$.

The worker $w$ computes a sub-partition of its local domain $\Sw$ in $M$ disjoint sub-domains $\Cmw$ of size $2^d|\Theta|$. Then, at each iteration $q$, the worker $w$ chooses an update candidate
\begin{equation}
    \label{eq:gcd_candidate}
    (k_0, \omega_0) = \argmax_{(k, \omega)\in[1, K]\times\Cmw}
        |\Delta Z_k[\omega]| \enspace ,
\end{equation}
\mbox{with $m = q \mod M$}. The process to accept or reject the update candidate uses a soft-lock mechanism described in the following paragraphs. If the candidate is accepted, the value of the coordinate $Z_{k_0}[\omega_0]$ is updated to $Z'_{k_0}[\omega_0]$, beta is updated with \autoref{eq:beta_update} and the neighboring workers $w'$ are notified if $\mathcal V(\omega_0) \cap \mathcal S_{w'} \neq \emptyset$. If the candidate is rejected, the worker moves on to the next sub-domain $\mathcal C_{m+1}^{w}$ without updating a coordinate.

\paragraph{Interferences.}
\label{par:dicodil:dicod:interferences}

When $W$ coordinates $(k_w, \omega_w)_{w=1}^W$ of $Z$ are updated simultaneously by respectively $\Delta Z_{k_w}[\omega_w]$, the updates might not be independent. The local version of $\beta$ used for the update does not account for the other updates. The cost difference resulting from these updates is denoted $\Delta E$, and the cost reduction induced by only one update $w$ is denoted $\Delta E_{k_w}[\omega_w]$. Simple computations, detailed in \autoref{prop:ii}, show that
\begin{align}
    \Delta& E =
        \sum_{i=1}^W\Delta E_{k_w}[\omega_w]
    \label{eq:interf}\\
&\hskip-1em -
    \sum_{w \neq w'}(\pmb D_{k_w} * \ttran{\pmb D_{k_{w'}}})[\omega_{w'} - \omega_w]
    \Delta Z_{k_w}[\omega_w] \Delta Z_{k_{w'}}[\omega_{w'}]
    ~,\nonumber
\end{align}
If for all $w$, all other updates $w'$ are such that if $\omega_{w'} \notin \mathcal V(\omega_w)$, then $(\pmb D_{k_w} * \ttran{\pmb D_{k_{w'}}})[\omega_{w'}-\omega_w] = 0$ and the updates can be considered to be sequential as the interference term is zero. When $\left|\left\{\omega_w \right\}_w \cap \mathcal V(\omega_0)\right| = I_0 > 1$, the interference term does not vanish. \citet{Moreau2018} show that if $I_0 < 3$, then, under mild assumption, the interference term can be controlled and DICOD converges. This is sufficient for 1D partitioning as this is always verified in this case. However, their analysis cannot be extended to $I_0 \geq 3$, limiting the partitioning of $\Omega$.

\paragraph{Soft Locks.}
\label{par:dicodil:dicod:soft_locks}

To avoid this limitation, we propose the soft-lock, a novel mechanism to avoid interfering updates of higher order. The idea is to avoid updating concurrent coordinates simultaneously. A classical tool in computer science to address such concurrency issue is to rely on a synchronization primitive, called \emph{lock}. This primitive can ensure that one worker enters a given block of code. In our case, it would be possible to use this primitive to only allow one worker to make an update on $\mathcal B_L(\Sw)$. This would avoid any interfering update. The drawback of this method is however that it prevents the algorithm to be asynchronous, as typical lock implementations rely on centralized communications. Instead, we propose to use an implicit locking mechanism which keeps our algorithm asynchronous. This idea is the following.

Additionally to the sub-domain $\Sw$, each worker also maintains the value of $Z$ and $\beta$ on the $\Theta$-extension $\mathcal E_L(\Sw)$ of its sub-domain, defined for $\Sw = \prod_{i=1}^d [l_i, u_i[$ as
\begin{equation}
    \mathcal E_L(\Sw) = \prod_{i=1}^d [l_i - L_i, l_i[ \cup [u_i, u_i + L_i[~.
\end{equation}
These two quantities can be maintained asynchronously by sending the same notifications on an extended border $\mathcal B_{2L}(\Sw)$ with twice the size of $\Theta$ in each direction. Then, at each iteration, the worker $w$ gets a candidate coordinate $(k_0, \omega_0)$ to update from \autoref{eq:gcd_candidate}. If $\omega_0 \not\in \mathcal B_L(\Sw)$, the coordinate is updated like in LGCD. When $\omega_0 \in \mathcal B_L(\Sw)$, the candidate is accepted if there is no better coordinate update in $\mathcal V(\omega_0) \cap \mathcal E_L(\Sw)$, \ie{} if
\begin{equation}
    \label{eq:soft_lock}
    |\Delta Z_{k_0}[\omega_0]| >
        \max_{k, \omega \in [1, K]\times \mathcal V(\omega_0) \cap \mathcal E_L(\Sw)}
            |\Delta Z_k[\omega]| \enspace .
\end{equation}
In case of equality, the preferred update is the one included in the sub-domain $\mathcal S_{w'}$ with the minimal index. Using this mechanism effectively prevents any interfering update. If two workers $w < w'$ have update candidates $\omega_0$ and $\omega'_0$ such that $\omega'_0 \in\mathcal V(\omega_0)$, then  with \autoref{eq:soft_lock}, only one of the two updates will be accepted -- either the largest one if $|\Delta Z_{k_0}[\omega_0]| \neq |\Delta Z_{k_0'}[\omega'_0]|$ or the one from $w$ if both updates have the same magnitude -- and the other will be dropped. This way the $Z$ updates can be done in an asynchronous way.

\paragraph{Speed-up analysis of \mDICOD}

Our algorithm \mDICOD{} has a sub-linear speed-up with the number of workers $W$. As each worker runs LGCD locally, the computational complexity of a local iteration in a worker $w$ is $\bO{2^dK|\Theta|}$. This complexity is independent of the number of workers used to run the algorithm. Thus, the speed-up obtained is equal to the number of updates that are performed in parallel. If we consider $W$ update candidates $(k_w, \omega_w)$ chosen by the workers, the number of updates that are performed corresponds to the number of updates that are not soft locked. When the updates are uniformly distributed on $\Sw$, the probability that an update candidate located in $\omega\in\Sw$ is not soft locked can be lower bounded by
\begin{equation}
    P(\omega \not\in \textbf{SL}) \ge \prod_{i=1}^d (1 - \frac{W_iL_i}{T_i})
\end{equation}
Computations are detailed in \autoref{prop:iii}. When the ratios $\frac{T_i}{W_iL_i}$ are large for all $i$, \ie{} when the size of the workers sub-domains $|\Sw|$ are large compared to the dictionary size $|\Theta|$, then $P(\omega \not\in \textbf{SL}) \simeq 1$. In this case, the expected number of accepted update candidates is close to $W$ and \mDICOD{} scales almost linearly. When $W_i$ reaches $\frac{T_i}{L_i} (\frac{2^{1/d}}{2^{1/d} - 1})$, then $ P(\omega \not\in \textbf{SL}) \gtrsim \frac{1}{2}$, and the acceleration is reduced to $\frac{W}{2}$.

In comparison, \citet{Moreau2018} reports a super-linear speed-up for DICOD on 1D signals. This is due to the fact that they are using GCD locally in each worker, which has a very high iteration complexity when $W$ is small, as it scales linearly with the size of the considered sub domain. Thus, when sub-dividing the signal between more workers, the iteration complexity is decreasing and more updates are performed in parallel, which explains why the speed-up is almost quadratic for low $W$. As the complexity of iterative GCD are worst than LGCD, \mDICOD{} has better performance than DICOD in low $W$ regime. Furthermore, in the 1D setting, DICOD becomes similar to \mDICOD{} once the size of the workers sub-domain becomes too small to be further partitioned with sub-segments $\Cmw$ of size $2^d|\Theta|$. The local iteration of \mDICOD{} are the same as the one performed by DICOD in this case. Thus, DICOD does not outperform \mDICOD{} in the large $W$ setting either.

\subsection{Distributed dictionary updates}
\label{sub:dicodil:sufficient}

Dictionary updates need to minimize a quadratic objective under constraint \autoref{eq:dict_update}. DiCoDiLe uses PGD to update its dictionary, with an Armijo backtracking line-search \citep{Wright1999}. Here, the bottle neck is that the computational complexity of the gradient for this objective is $\bO{PK^2|\Omega|\log(|\Omega|)}$. For very large signals or images, this cost is prohibitive. We propose to also use the distributed set of workers to scale the computations. The function to minimize in \autoref{eq:dict_update} can be factorized as
\begin{equation}
    \nabla_{\pmb D} F(Z, \pmb D) = \ttran{Z} * ( X - Z * \pmb D) = \psi -  \phi*\pmb D \enspace ,
\end{equation}
with $\phi\in\Xset_\Phi^{K\times K}$ being the restriction of the convolution $\ttran{Z}*Z$ to $\Phi = \prod_{i=1}^d [-L_i + 1, L_i[$, and $\psi \in \Xset_\Theta^{K\times P}$ the convolution $\ttran{Z}*X$ restricted to $\Theta$. These two quantities can be computed in parallel by the workers. For $\tau \in \Phi$,
\begin{equation}
    \label{eq:phi}
    \phi[\tau] = \sum_{\omega\in\Omega} Z[\omega]Z[\tau + \omega]
               = \sum_{w=1}^W \sum_{\omega\in\Sw} Z[\omega]Z[\tau + \omega]
\end{equation}
and for all $\tau, \omega \in \Phi\times\Sw$, we have $\omega - \tau \in \Sw \cup \mathcal E_\Theta(\Sw)$. Thus, the computations of the local $\phi^w$ can be mapped to each worker, before being provided to the reduce operation $\phi = \sum_{w=1}^W \phi^w$. The same remark is valid for the computation of $\ttran{Z} * X$. By making use of these distributed computation, $\phi$ and $\psi$ can be computed with respectively $\bO{K^2|\Sw|\log(|\Sw \cup \mathcal E_\Theta(\Sw)|)}$ and $\bO{KP|\Sw|\log(|\Sw \cup \mathcal E_\Theta(\Sw)|)}$ operations.
The values of $\psi$ and $\phi$ are sufficient statistics to compute $\nabla_{\pmb D} F$ and $F$, which can thus be computed with complexity $\bO{K^2P|\Theta|\log(2^d|\Theta|)}$, independently of the signal size. This allows to efficiently update the dictionary atoms.

\section{Numerical Experiments}
\label{sec:expes}

The numerical experiments are run on a SLURM cluster with 30 nodes. Each node has 20 physical cores, 40 logical cores and 250\,GB of RAM. DiCoDiLe is implemented in Python \citep{Python36} using \texttt{mpi4py} \citep{Dalcin2005}\footnote{Code available at \href{https://github.com/tomMoral/dicodile}{github.com/tommoral/dicodile}}.

\begin{figure}[tb]
    \begin{center}
        \centerline{\includegraphics[width=0.7\columnwidth]{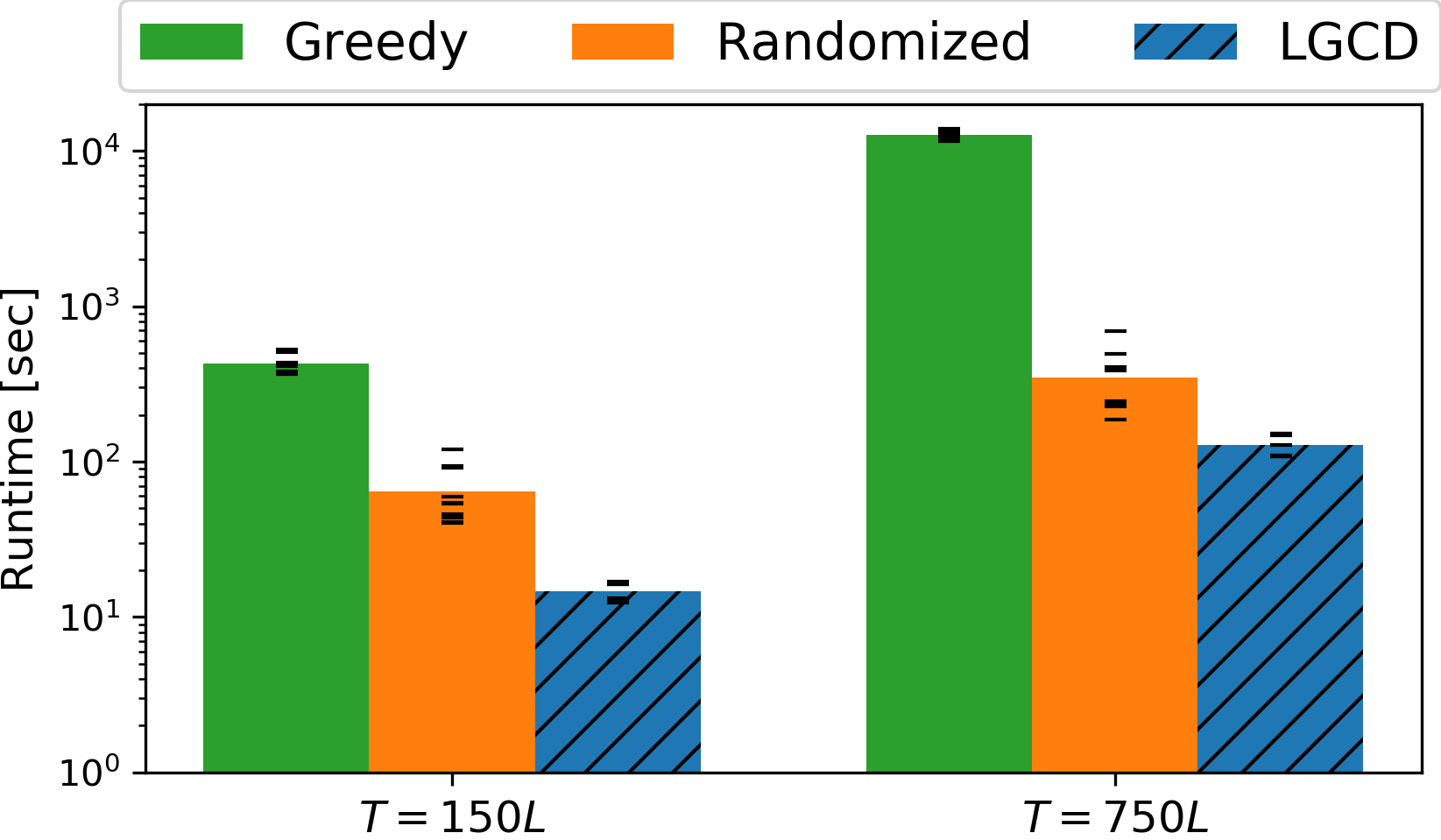}}
        \caption{Average running time of CD with three coordinate selection schemes -- (\emph{blue}) Locally Greedy, (\emph{orange}) Randomized and (\emph{green}) Greedy -- for two signal lengths. LGCD consistently outperforms the two other strategies.}
        \label{fig:compare_cd}
    \end{center}
\end{figure}

\subsection{Performance of distributed sparse coding}

The distributed sparse coding step in DiCoDiLe is impacted by
the coordinate descent strategy, as well as the data partitioning and
soft-lock mechanism. The following experiments show their impact on \mDICOD.

\paragraph{Coordinate Selection Strategy}
\label{par:expes:strategy}

\begin{figure}[tb]
    \begin{center}
        \centerline{\includegraphics[width=.7\columnwidth]{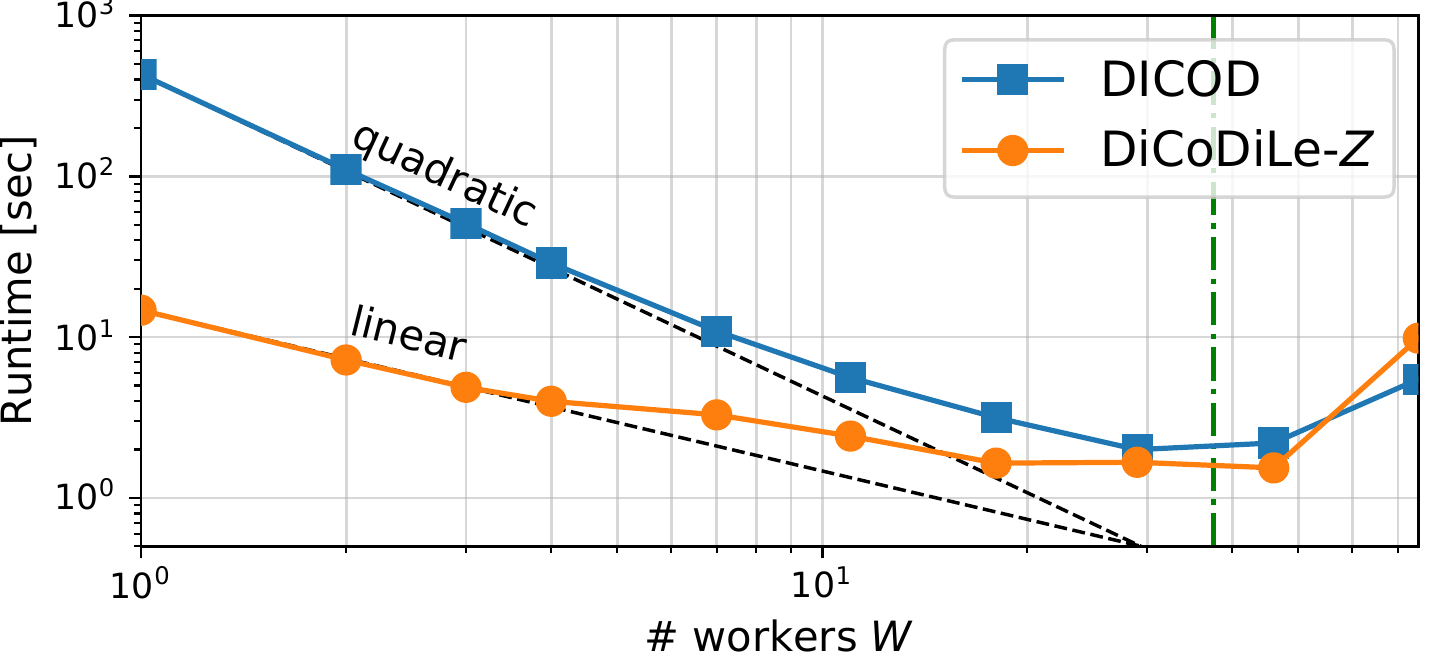}}
        \caption{Scaling of DICOD (\emph{orange}) and \mDICOD{} (\emph{blue}) with the number of workers $W$. \mDICOD{} only scales sub-linearly while DICOD scales super-linearly. However \mDICOD{} is more efficient that DICOD as the performance of the former with one worker is quite poor. When $W$ reaches $T / 4L$ (\emph{dashed-green}), \mDICOD{} and DICOD become equivalent as \mDICOD{} only has one sub-domain in each worker.}
        \label{fig:scaling_1d}
    \end{center}
\end{figure}

The experimental data are generated following the sparse convolutional linear model \autoref{eq:model} with $d=1$ in $\Rset^P$ with $P=7$. The dictionary is composed of $K=25$ atoms $\pmb D_k$ of length $L=250$. Each atom is sampled from a standard Gaussian distribution and then normalized. The sparse code entries are drawn from a Bernoulli-Gaussian distribution, with Bernoulli parameter $\rho = 0.007$, mean $0$ and standard variation $10$. The noise term $\xi$ is sampled from a standard Gaussian distribution with variance 1. The length of the signals $X$ is denoted $T$. The regularization parameter was set to $0.1\lambda_{\max}$.

Using one worker, \autoref{fig:compare_cd} compares the average running times of three CD strategies to solve \autoref{eq:csc}: Greedy (GCD), Randomized (RCD) and LGCD. LGCD consistently outperforms the two other strategies for both $T=150L$ and $T=750L$. For both LGCD and RCD, the iteration complexity is constant compared to the size of the signal, whereas the complexity of each GCD iteration scales linearly with T. This makes running time of GCD explode when $T$ grows. Besides, as LGCD also prioritizes the more important coordinates, it is more efficient than uniform RCD. This explains \autoref{fig:compare_cd} and justifies the use of LGCD is the following experiments.

\autoref{fig:scaling_1d} investigates the role of CD strategies in the distributed setting.
The average runtimes of DICOD (with GCD on each worker), and \mDICOD{} (with LGCD) are reported in \autoref{fig:scaling_1d}. We stick to a small scale problem in 1D to better highlight the limitation of the algorithms but scaling results for a larger and 2D problems are included in \autoref{fig:supp:scaling_1d} and \autoref{fig:annex:scaling_2d}. \mDICOD{} with LGCD scales sub-linearly with the number of workers used, whereas DICOD scales almost quadratically. However, \mDICOD{} outperforms DICOD consistently, as the performance of GCD is poor for low number of workers using large sub-domains. The two coordinate selection strategies become equivalent when $W$ reaches $T/4L$, as in this case DiCoDiLe has only one sub-domain $\mathcal C^{(w)}_1$.

\paragraph{Impact of 2D grid partitioning on images}
\label{par:expes:soft_lock}

The performance of \mDICOD{} on images are evaluated using the standard colored image Mandrill at full resolution ($512\times 512$). A dictionary $\pmb D$ consists of $K=25$ patches of size $16\times 16$ extracted from the original image. We used $\lambda = 0.1\lambda_{\max}$ as a regularization parameter.

\begin{figure}[tb]
    \begin{minipage}{\columnwidth}
        \begin{minipage}{0.48\columnwidth}
            \includegraphics[width=.8\columnwidth]{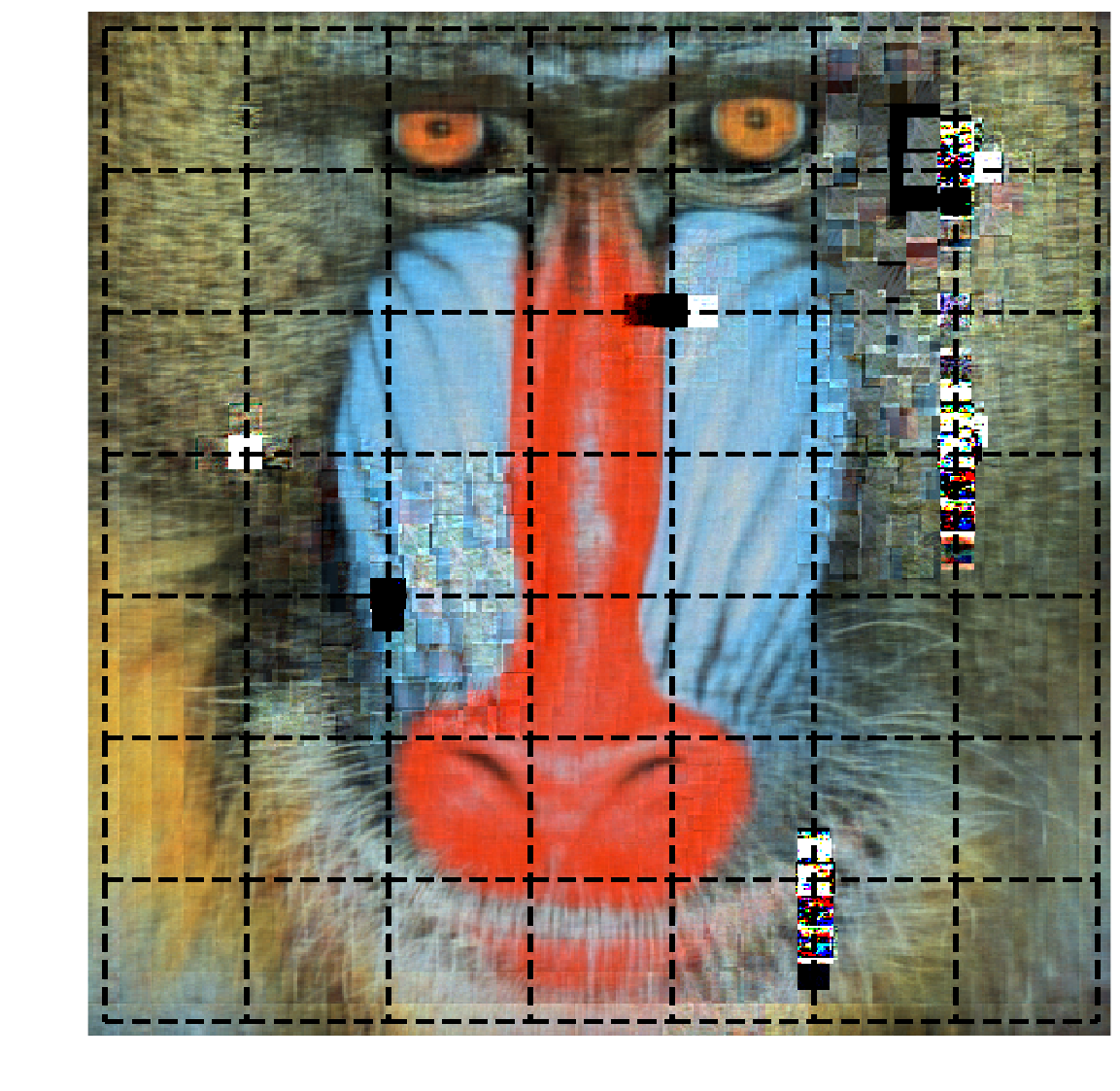}
        \end{minipage}
        \hfill
        \begin{minipage}{0.49\columnwidth}
            \vspace{-0.18in}
            \caption{Reconstruction of the image Mandrill using DiCoDiLe with \textbf{no} soft-locks and $49$
                     workers. The dashed lines show the partitions of the domain $\Omega$. The algorithm diverges
                     at the edges of some of the sub-domains due to interfering updates between more than two
                     workers.}
            \label{fig:soft_lock}
        \end{minipage}
    \end{minipage}
\end{figure}

\begin{figure}[tb]

\begin{center}
    \centerline{\includegraphics[width=.7\columnwidth]{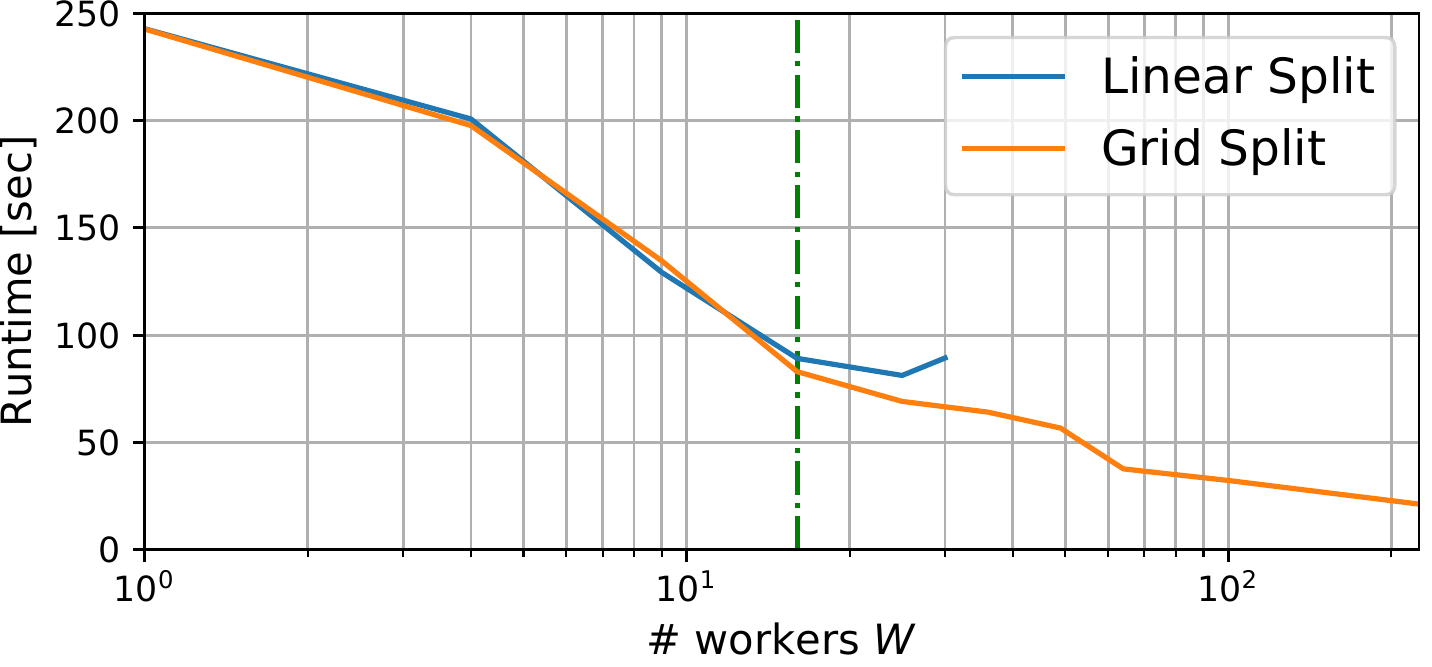}}
    \caption{Scaling on 2D images of \mDICOD{} with the number of workers for two partitioning strategies of $\Omega$:
             (\emph{blue}) $\Omega$ is split only along one direction, as in DICOD,
             (\emph{orange}) $\Omega$ is split along both directions, on a grid.
             The running times are similar for low number of workers
             but the performances with the linear splits stop improving once $W$ reaches the scaling limit $T_1/4L_1$ (\emph{green}). With the grid of workers adapted to images, the performance improves further. The linear split stops when $W$ reached $T_1/L_1$ because no more coordinate in $\Sw$ are independent from other neighbors.}
    \label{fig:scaling_grid}
\end{center}
\end{figure}

The reconstruction of the image with no soft-locks is shown in \autoref{fig:soft_lock}. To ensure the algorithm finishes, workers are stopped either if they reach convergence or if $\|Z\|_\infty$ becomes larger than $\min_k \frac{50}{\|\pmb D_k\|_\infty}$ locally. When a coefficient reaches this value, at least one pixel in the encoded patch has a value 50 larger than the maximal value for a pixel, so we can safely assume algorithm is diverging. The reconstructed image shows artifacts around the edges of some of the sub-domains $\Sw$, as the activation are being wrongly estimated in these locations. This visually demonstrates the need for controlling the interfering updates when more than two workers might interfere. When using \mDICOD{}, the workers do not diverge anymore due to soft-locks.

To demonstrate the impact of the domain partitioning strategy on images, \autoref{fig:scaling_grid} shows the average runtime for \mDICOD{} using either a line of worker or a grid of workers. In this experiment $K=5$ and the atom size is $8\times 8$. Both strategies scale similarly in the regime with low number of workers. But when $W$ reaches $T_1/3L_1$, the performances of \mDICOD{} stops improving when using the unidirectional partitioning of $\Omega$. Indeed in this case many update candidates are selected at the border of the sub-domains $\Sw$, therefore increasing the chance to reject an update. Moreover, the scaling of the linear split is limited to $W = T_1/2L_1 = 32$, whereas the 2D grid split can be used with $W$ up to $1024$ workers.

\subsection{Application to real signals}

\paragraph{Learning dictionary on Hubble Telescope images}
\label{par:expes:astro}

\begin{figure}[t!]

    \begin{center}
        \centerline{\includegraphics[width=.9\columnwidth]{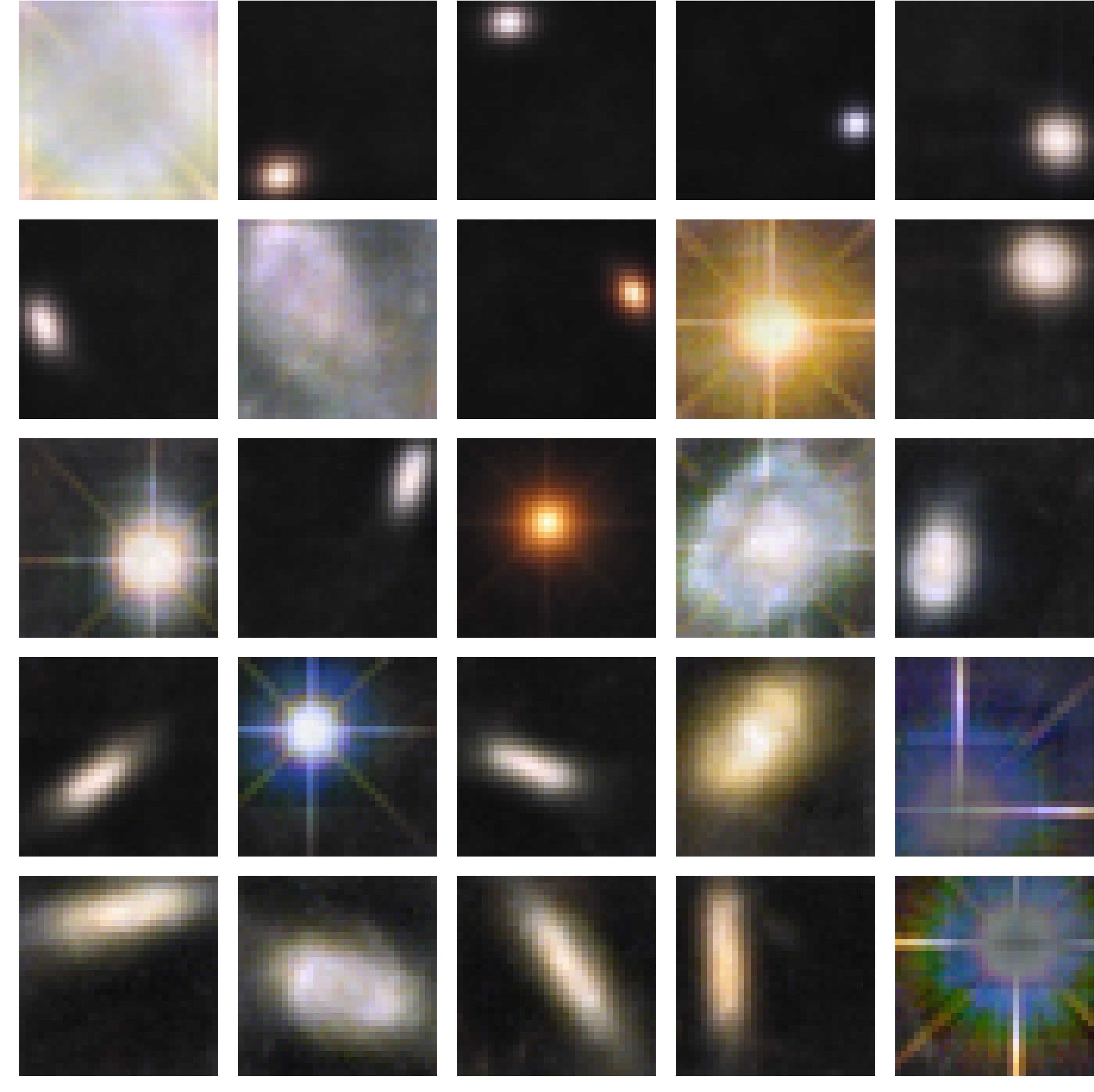}}
        \caption{ $25$ atoms $32\times 32$  learned from the Hubble Telescope images (\texttt{STScI-H-2016-39-a}) using
            DiCoDiLe with $400$ workers and a regularization parameter $\lambda=0.1\lambda_{\max}$. The atoms are
            sorted based on $\|Z_k\|_1$ from the top-left to the bottom-right. Most atoms
            displays a structure which corresponds to spatial objects. The atoms 1 and 7
            have fuzzier structure because they are mainly used to encode larger scale objects.}
        \label{fig:astro_dict}
    \end{center}
\end{figure}

We present here results using CDL to learn patterns from an image of the Great Observatories Origins Deep Survey (GOODS) South field, acquired by the Hubble Space Telescope \citep{Giavalisco2004}. We used the STScI-H-2016-39 image\footnote{The image is at \href{http://hubblesite.org/image/3920/news/58-hubble-ultra-deep-field}{www.hubblesite.org/image/3920/news/58-hubble-ultra-deep-field}} with resolution $6000\times 3600$, and used DiCoDiLe to learn 25 atoms of size $32\times 32$ with regularization parameter $\lambda = 0.1\lambda_{\max}$. The learned patterns are presented in \autoref{fig:astro_dict}. This unsupervised algorithm is able to highlight structured patterns in the image, such as small stars. Patterns 1 and 7 are very fuzzy. This is probably due to the presence of very large object in the foreground. As this algorithm is not scale invariant, the objects larger than our patterns are encoded using these low frequency atoms.

For comparison purposes, we run the parallel Consensus ADMM algorithm\footnote{Code available at \href{https://sporco.readthedocs.io/en/latest/}{https://sporco.readthedocs.io/}} proposed by \citet{Skau2018} on the same data with one node of our cluster but it failed to fit the computations in 250\,GB of RAM. We restricted our problem to a smaller patch of size $512\times512$ extracted randomly in the Hubble image. \autoref{fig:annex:compare_cdl} in supplementary material reports the evolution of the cost as a function of the time for both methods using $W=36$. The other algorithm performs poorly in comparison to DiCoDiLe.

\FloatBarrier

\section{Conclusion}
\label{sec:conclusion}

This work introduces DiCoDiLe, an asynchronous distributed algorithm to solve the CDL for very large multidimensional data such as signals and images. The number of workers that can be used to solve the CSC scales linearly with the data size, allowing to adapt the computational resources to the problem dimensions. Using smart distributed pre-computation, the complexity of the dictionary update step is independent of the data size. Experiments show that it has good scaling properties and demonstrate that DiCoDiLe is able to learn patterns in astronomical images for large size which are not handled by other state-of-the-art parallel algorithms.

\bibliographystyle{plainnat}
\bibliography{library}

\newpage

\appendix

\setcounter{figure}{0}
\renewcommand{\thefigure}{\Alph{section}.\arabic{figure}}
\setcounter{equation}{0}
\renewcommand{\theequation}{\Alph{section}.\arabic{equation}}

\section{Computation for the cost updates}
\label{sec:annex:cost_update}

When a coordinate $Z_k[t]$ is updated to $Z'_k[t]$, the cost update is a simple function of $Z_k[t]$ and $Z'_k[t]$.

\begin{propositionA}
    If the coordinate $(k_0, \omega_0)$ is updated from value $Z_{k_0}[\omega_0]$ to value $Z'_{k_0}[\omega_0]$, then the change in the cost function $\Delta E_{k_0}[\omega_0]$ is given by:
    \begin{align*}
    \Delta E_{k_0}[\omega_0] & = \frac{\|\pmb D_{k_0}\|_2^2}{2} (Z_{k_0}[\omega_0]^{2} - Z'_{k_0}[\omega_0]^2)
    - \beta_{k_0}[\omega_0](Z_{k_0}[\omega_0] - Z'_{k_0}[\omega_0])
    + \lambda(|Z_{k_0}[\omega_0]| - |Z'_{k_0}[\omega_0]|).
    \end{align*}%
    \label{propA:pb_coord}
\end{propositionA}

\begin{proof}

    We denote $Z^{(1)}_{k}[\omega] =
    \begin{cases}
    Z'_{k_0}[\omega_0], &\text{ if } (k, \omega) = (k_0, \omega_0)\\
    Z_{k}[\omega], &\text{ elsewhere }\\
    \end{cases}$. Let $\alpha_{0}$ be the residual when coordinate $(k_0, \omega_0)$ of $Z$ is set to 0.
    For $\omega \in \Omega$,
    \[
    \alpha_{k_0}[\omega] = (X - Z*\pmb D)[\omega] + \pmb D_{k_0}[\omega-\omega_0] Z_{k_0}[\omega_0]
    = (X - Z*\pmb D)[\omega] + Z_{k_0}[\omega_0](e_{\omega_0}*\pmb D_{k_0})[\omega_0]~.
    \]
    We also have $\alpha_{k_0}[\omega] = (X - Z^{(1)}*\pmb D)[\omega] + \pmb D_{k_0}[\omega-\omega_0] Z'_{k_0}[\omega_0]$.
    Then
    \begin{align*}
    \Delta E_{k_0}[\omega_0] =& \frac{1}{2}\sum_{\omega\in\Omega} \left(X - Z*\pmb D\right)^2[\omega] + \lambda \|Z\|_1
    - \frac{1}{2}\sum_{\omega\in\Omega} \left(X - Z^{(1)}*\pmb D\right)^2[\omega] + \lambda \|Z^{(1)}\|_1 \\
    =& \frac{1}{2}\sum_{\omega\in\Omega}\left(\alpha_{k_0}[\omega] - \pmb D_{k_0}[\omega-\omega_0] Z_{k_0}[\omega_0]\right)^2
    - \frac{1}{2}\sum_{\omega\in\Omega} \left(\alpha_{k_0}[\omega] - \pmb D_{k_0}[\omega-\omega_0] Z'_{k_0}[\omega_0]\right)^2
    + \lambda (|Z_{k_0}[\omega_0]| - |Z'_{k_0}[\omega_0]|)\\
    =& \frac{1}{2}\sum_{\omega\in\Omega} \pmb D_{k_0}[\omega-\omega_0]^2(Z_{k_0}[\omega_0]^2 - Z'_{k_0}[\omega_0]^2)
    - \sum_{\omega\in\Omega}\alpha_{k_0}[\omega]\pmb D_{k_0}[\omega-\omega_0] (Z_{k_0}[\omega_0] - Z'_{k_0}[\omega_0])
    + \lambda (|Z_{k_0}[\omega_0]| - |Z'_{k_0}[\omega_0]|)\\
    =& \frac{\|\pmb D_{k_0}\|_2^2}{2} (Z_{k_0}[\omega_0]^2 - Z'_{k_0}[\omega_0]^2)
    - \underbrace{(\alpha_{k_0} * \ttran{\pmb D_{k_0}})[\omega_0]}_{\beta_{k_0}[\omega_0]}(Z_{k_0}[\omega_0] - Z'_{k_0}[\omega_0])
    + \lambda (|Z_{k_0}[\omega_0]| - |Z'_{k_0}[\omega_0]|)
    \end{align*}
    This conclude our proof.
\end{proof}

Using this result, we can derive the optimal value $Z'_{k_0}[\omega_0]$ to update the coordinate $(k_0, \omega_0)$ as the solution of the following optimization problem:
\begin{equation}
    \label{eqA:pb_coord}
    Z'_{k_0}[\omega_0] = \arg\min_{y\in \Rset} e_{k_0, \omega_0}(u) \sim \arg\min_{u\in \Rset}\frac{\| \pmb D_{k_0}\|_2^2}{2}\left(u - \beta_{k_0}[\omega_0]\right)^2 + \lambda |u|~.\\
\end{equation}
In the case where multiple coordinates $(k_w, \omega_w)$ are updated in the same iteration
to values $Z'_{k_w}[\omega_w]$, we obtain the following cost variation.

\begin{propositionA}\label{prop:ii}
    The update of the $W$ coordinates $(k_w, \omega_w)_{w=1}^W$ with additive update $\Delta Z_{k_w}[\omega_w]$ changes the cost by:
    \begin{align*}
    \Delta E =
    \overbrace{\sum_{i=1}^W\Delta E_w}^{
        \text{iterative steps}}
    - \underbrace{\sum_{w \neq w'}(\pmb D_{k_w} * \ttran{\pmb D_{k_{w'}}})[\omega_{w'} - \omega_w]
        \Delta Z_{k_w}[\omega_w] \Delta Z_{k_{w'}}[\omega_{w'}]}_{
        \text{interference}}, \label{eq:interf}
    \end{align*}
\end{propositionA}

\begin{proof}
    We define $Z_k^{(1)}[t] = \begin{cases}
    Z_{k_w}[\omega_w], &\text{ if } (k, \omega) = (k_w, \omega_w)\\
    Z_k[t], &\text{ otherwise }
    \end{cases}~.$\\[.3em]

    Let
    \[
    \alpha[\omega] = (X - Z*\pmb D)[t] + \sum_{w=1}^K\pmb D_{k_w}[\omega-\omega_w] Z_{k_w}[\omega_w]
    =  (X - Z^{(1)}*\pmb D)[t] + \sum_{w=1}^K\pmb D_{k_w}[\omega-\omega_w] Z'_{k_w}[\omega_w]~,
    \]
    and
    \[\alpha_{w}[\omega] = (X - Z*\pmb D)[t] + \pmb D_{k_w}[\omega-\omega_w] Z_{k_w}[\omega_w] = \alpha[\omega] - \sum_{w'\neq w} \pmb D_{k_{w'}}[\omega-\omega_{w'}] Z_{k_{w'}}[\omega_{w'}]~.
    \]
    Then
    \begin{align*}
    \Delta E =
    & \frac{1}{2}\sum_{\omega\in\Omega} \left(X[\omega] - Z*\pmb D[\omega]\right)^2
    + \lambda \|Z\|_1
    - \sum_{\omega\in\Omega} \left(X[\omega] - Z^{(1)}*\pmb D[\omega]\right)^2
    + \lambda \|Z^{(1)}\|_1 \\
    =
    & \frac{1}{2}\sum_{\omega\in\Omega}\left(\alpha[\omega]
    - \sum_{w=1}^W\pmb D_{k_w}[\omega-\omega_w] Z_{k_w}[\omega_w]\right)^2
    - \frac{1}{2}\sum_{\omega\in\Omega} \left(\alpha[\omega]
    - \sum_{w=1}^W\pmb D_{k_w}[\omega-\omega_w] Z'_{k_w}[\omega_w]\right)^2 \\
    & + \sum_{w=1}^W\lambda (|Z_{k_w}[\omega_w]| - |Z'_{k_w}[\omega_w]|)\\
    =
    &\frac{1}{2}\sum_{\omega\in\Omega} \sum_{w=1}^W \pmb D_{k_w}[\omega - \omega_w]^2
    (Z_{k_w}[\omega_w]^2 - {Z'_{k_w}[\omega_w]}^2)
    + \sum_{w=1}^W\lambda (|Z_{k_w}[\omega_w]| - |Z'_{k_w}[\omega_w]|)\\
    & -\sum_{\omega\in\Omega}\Bigl[\sum_{w=1}^W\alpha[\omega] \pmb D_{k_w}[\omega-\omega_w]
    \Delta Z_{k_w}[\omega_w]\\
    &~~~~~~~~~~~~~~~
    - \sum_{w \neq w'} \pmb D_{k_w}[\omega-\omega_w] \pmb D_{k_{w'}}[\omega-\omega_{w'}]
    (Z_{k_w}[\omega_w]Z_{k_{w'}}[\omega_{w'}] - Z'_{k_w}[\omega_w]Z'_{k_{w'}}[\omega_{w'}])\Bigr]\\
    =
    & \sum_{w=1}^W(\|\pmb D_{k_w}\|_2^2(Z_{k_w}[\omega_w]^2-{Z'_{k_w}[\omega_w]}^2)
    - (\alpha_{k_w}*\ttran{\pmb D_{k_w}})[\omega_w]\Delta Z_{k_w}[\omega_w]
    + \lambda (|Z_{k_w}[\omega_w]| - |Z'_{k_w}[\omega_w]|)\\
    & - \sum_{\omega\in\Omega}\Bigl[
    \sum_{w=1}^W\alpha_{k_w}[\omega] \pmb D_{k_w}[\omega-\omega_w]\Delta Z_{k_w}[\omega_w]\\
    &~~~~~~~~~~~~~~ +
    \sum_{w \neq w'}  \pmb D_{k_w}[\omega-\omega_w] \pmb D_{k_{w'}}[\omega-\omega_{w'}]
    (\Delta Z_{k_w}[\omega_w] Z'_{k_{w'}}[\omega_{w'}] +
    \Delta Z_{k_{w'}}[\omega_{w'}] Z'_{k_w}[\omega_w])\\
    &~~~~~~~~~~~~~~ ~~~~~~~~~~
    - \pmb D_{k_w}[\omega-\omega_w] \pmb D_{k_{w'}}[\omega-\omega_{w'}]
    (Z_{k_w}[\omega_w]Z_{k_{w'}}[\omega_{w'}] -
    Z'_{k_w}[\omega_w]Z'_{k_{w'}}[\omega_{w'}])\Bigr]\\
    =
    & \sum_{w=1}^W \Delta E_{k_w}[\omega_w] - \sum_{w \neq w'}\Biggl[\Bigl(
    \sum_{\omega\in\Omega}  \pmb D_{k_w}[\omega-\omega_w]
    \pmb D_{k_{w'}}[\omega-\omega_{w'}]\Bigr) \times\\
    & \hskip6em\Bigl[
    Z_{k_w}[\omega_w]Z_{k_{w'}}[\omega_{w'}] - Z'_{k_w}[\omega_w]Z_{k_{w'}}[\omega_{w'}]
    - Z_{k_w}[\omega_w]Z'_{k_{w'}}[\omega_{w'}] + Z'_{k_{w'}}[\omega_{w'}]Z'_{k_w}[\omega_w]
    \Bigr]\Biggr]\\
    =
    & \sum_{w=1}^W\Delta E_{k_w}[\omega_w]
    - \sum_{w \neq w'} \left( \sum_{\omega\in\Omega} \pmb D_{k_w}[\omega]
    \pmb D_{k_{w'}}[\omega+\omega_w-\omega_{w'}]\right)
    (Z_{k_w}[\omega_w] - Z'_{k_w}[\omega_w])(Z_{k_{w'}}[\omega_1] - Z'_{k_{w'}}[\omega_{w'}]) \\
    =
    & \sum_{w=1}^W\Delta E_{k_w}[\omega_w]
    - \sum_{w \neq w'} (\pmb D_{k_{w'}} * \ttran{\pmb D_{k_w}})[\omega_w-\omega_{w'}]
    \Delta Z_{k_w}[\omega_w] \Delta Z_{k_{w'}}[\omega_{w'}]
    \end{align*}
\end{proof}

\section{Probability of a coordinate to be soft-locked}
\label{sec:annex:proba_interf}

\setcounter{equation}{0}

\begin{propositionA}
    \label{prop:iii}
    The probability of an update candidate being accepted, if the updates are spread uniformly
    on $\Omega$, is lower bounded by
    \[
    P(\omega \not\in \textbf{SL}) \ge \prod_{i=1}^d (1 - \frac{W_iL_i}{T_i})
    \]

\end{propositionA}

\begin{proof}
    Let $w \in [1, W]$ be a worker indice and $m \in [1, M]$ be the indice of the current sub-domain being considerd by the worker $w$. We denote $\omega = \argmax_{\omega \in \Cmw} \max_k |\Delta Z_k[\omega]|$ the position of the update candidate chosen by $w$ and $j = |\{w ~\emph{ s.t. }~ \Sw \cap \mathcal V(\omega) = \emptyset\}|$ the number of workers that are impacted by this update.
    With the definition of the soft-lock, there exists at least one worker in $\{w ~\emph{ s.t. }~ \Sw \cap \mathcal V(\omega) = \emptyset\}$ has a coordinate which is not soft-locked. As the updates are uniformly spread across the workers, we get the probability of $\omega$ being locked is $P(\omega \not\in \textbf{SL}~|~\omega) \ge \frac{1}{j}$.

    Thus, the probability of an update candidate $(k, \omega) \in [1, K]\times\Sw$ to be accepted is
    \begin{align}
    P(\omega \not\in \textbf{SL}) & = \sum_{\omega\in\Sw} P(\omega \not\in \textbf{SL}~|~\omega) P(\omega)\\
    & \ge \frac{1}{|\Sw|}\sum_{\omega\in\Sw} \frac{1}{j_\omega} \\
    & \ge \frac{1}{|\Sw|}\prod_{i=1}^d (T_i - W_iL_i)\label{eq:proof:count}\\
    & \ge \prod_{i=1}^d (1 - \frac{W_iL_i}{T_i})
    \end{align}
    where \autoref{eq:proof:count} results from the summation on $\Omega$. This step is clearer by looking at \autoref{fig:proof:count}.

    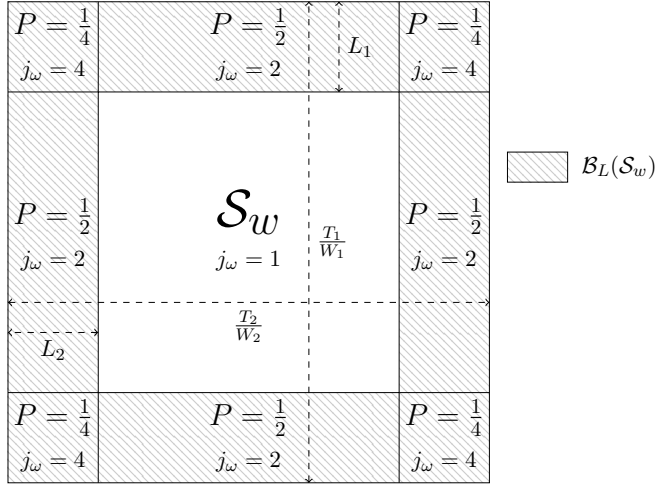
\begin{figure}[h!]
        \begin{minipage}{0.68\columnwidth}
            \scalebox{.8}{

        \begin{tikzpicture}

        \pgfmathsetmacro{\lin}{2.5}
        \pgfmathsetmacro{\lout}{4}
        \pgfmathsetmacro{\lmiddle}{3.25}

        \draw [pattern=north west lines, pattern color=gray!40] (-\lout, \lout) -- (-\lout, -\lout) -- (\lout, -\lout) -- (\lout, \lout) -- cycle;
        \draw [fill=white] (-\lin, \lin) -- (-\lin, -\lin) -- (\lin, -\lin) -- (\lin, \lin) -- cycle;

        \draw (-\lin, \lin) -- (-\lin, \lout);
        \draw (-\lin, \lin) -- (-\lout, \lin);
        \draw (\lin, \lin) -- (\lin, \lout);
        \draw (\lin, \lin) -- (\lout, \lin);
        \draw (\lin, -\lin) -- (\lin, -\lout);
        \draw (\lin, -\lin) -- (\lout, -\lin);
        \draw (-\lin, -\lin) -- (-\lin, -\lout);
        \draw (-\lin, -\lin) -- (-\lout, -\lin);

        \draw (0, 0) node[above] {\Huge$\mathcal S_w$} node [below] {$j_\omega = 1$};
        \draw  (\lmiddle, \lmiddle - .1)  node [above] {\Large $P=\frac{1}{4}$} node [below] {$j_\omega = 4$};
        \draw  (-\lmiddle, \lmiddle - .1)  node [above] {\Large $P=\frac{1}{4}$} node [below] {$j_\omega = 4$};
        \draw  (-\lmiddle, -\lmiddle - .1)  node [above] {\Large $P=\frac{1}{4}$} node [below] {$j_\omega = 4$};
        \draw  (\lmiddle, -\lmiddle- .1)  node [above] {\Large $P=\frac{1}{4}$} node [below] {$j_\omega = 4$};
        \draw  (0, -\lmiddle - . 1)  node [above] {\Large $P=\frac{1}{2}$} node [below] {$j_\omega = 2$};
        \draw  (0, \lmiddle - .1)   node [above] {\Large $P=\frac{1}{2}$} node [below] {$j_\omega = 2$};
        \draw  (-\lmiddle, 0)   node [above] {\Large $P=\frac{1}{2}$} node [below] {$j_\omega = 2$};
        \draw  (\lmiddle, 0)   node [above] {\Large $P=\frac{1}{2}$} node [below] {$j_\omega = 2$};
        \draw [<->, dashed] (-\lin, -1.5) --  node [below] {$L_2$} (-\lout, -1.5);
        \draw [<->, dashed] (1.5, \lin) --  node [right] {$L_1$} (1.5, \lout);
        \draw [<->, dashed] (1, -\lout) --  node [right] {$\frac{T_1}{W_1}$} (1, \lout);
        \draw [<->, dashed] (-\lout, -1) --  node [below] {$\frac{T_2}{W_2}$} (\lout, -1);

        \draw [pattern=north west lines, pattern color=gray!40] (\lout + .3, 1) rectangle  (\lout + 1.3, 1.5);
        \draw ( \lout + 1.4, 1.25) node [right] {$\mathcal B_L(\mathcal S_w)$};

        \end{tikzpicture}

        }
        \end{minipage}
        \begin{minipage}{0.3\columnwidth}
            \vspace{-0.18in}
            \caption{Summary of the value of $j_\omega$ on a 2D example. We can easily see that the sum along
                each direction is $\frac{T_i}{W_i} - L_i$ on each workers. When multiplied by the number
                of workers $W_i$ along each direction, we obtain \autoref{eq:proof:count}.}
            \label{fig:proof:count}
        \end{minipage}
    \end{figure}

\end{proof}

\pagebreak

\section{Extra experiments}
\label{sec:annex:figures}
\setcounter{figure}{0}

\begin{figure}[ht]
    \begin{center}
        \centerline{
            \includegraphics[width=.5\columnwidth]{scaling_T150}
            \includegraphics[width=.5\columnwidth]{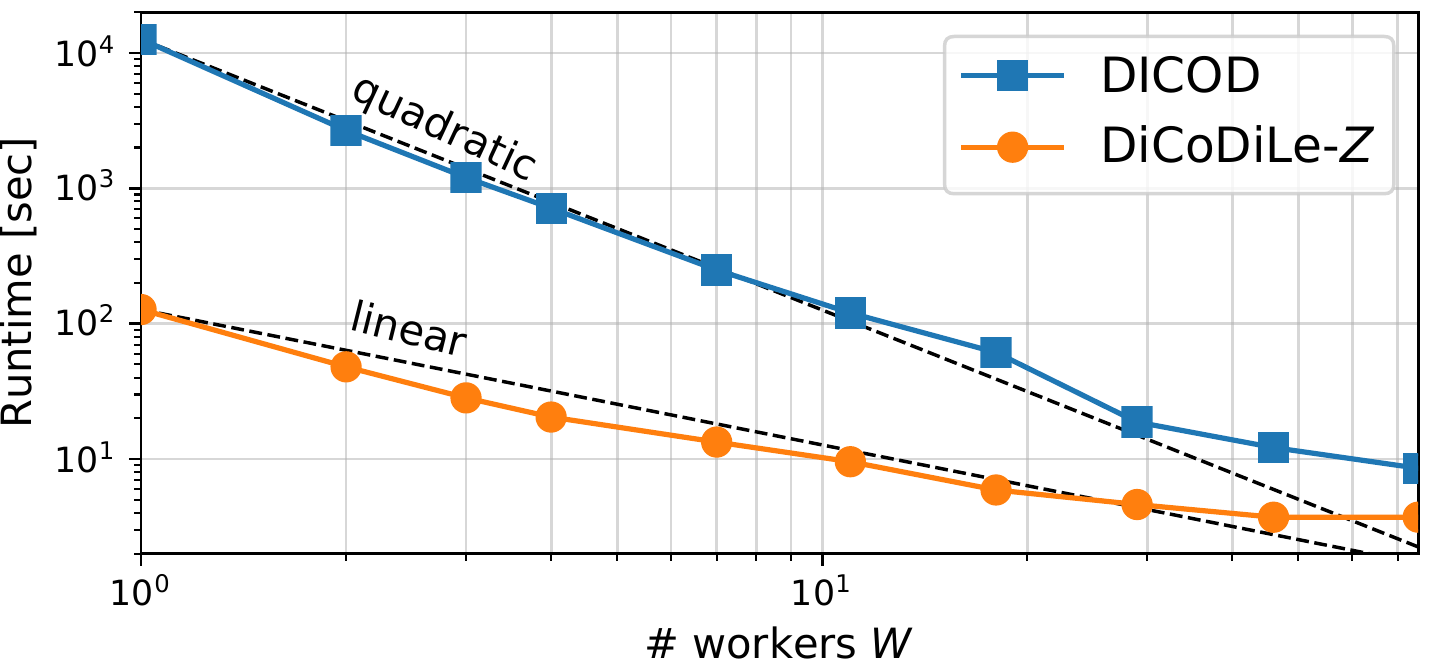}
        }
        \caption{Scaling of DICOD (\emph{orange}) and \mDICOD (\emph{blue}) with the number of workers $W$.
            \mDICOD only scales sub-linearly while DICOD scales super-linearly. However \mDICOD is more
            efficient that DICOD as the performance of the former with one worker are really bad. The
            green line denotes the number of cores where \mDICOD and DICOD become the same as \mDICOD
            only has one sub-domain in each worker. The results are consistent for both small
            (\emph{left}, $T=150L$) and large (\emph{right} $T=750L$) signals sizes.
        }
        \label{fig:supp:scaling_1d}
    \end{center}
\end{figure}

\paragraph{Scaling of \mDICOD{} for larger 1D signals} \autoref{fig:supp:scaling_1d} scaling properties of DICOD
and \mDICOD{} are consistent for larger signals. Indeed, in the case where $T=750$, \mDICOD{} scales linearly
with the number of workers. We do not see the drop in performance, as in this experiments $W$ does not reach the
scaling limit $T/3L$. Also, \mDICOD{} still outperforms DICOD for all regimes.

\begin{figure}[t]
    \centerline{\includegraphics[width=0.7\columnwidth]{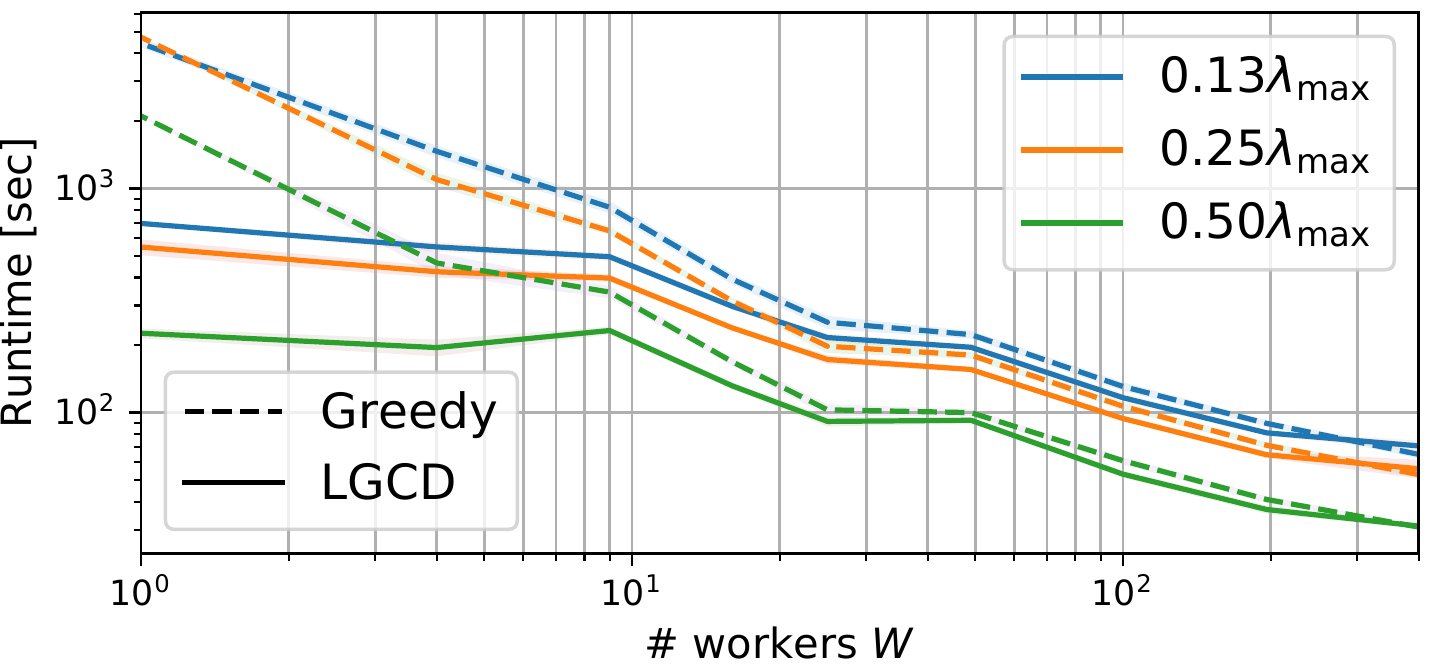}}
    \caption{
        Scaling of \mDICOD{} with the number of workers $W$ for different value of $\lambda$ and for two strategies of coordinate
        selection: Greedy and Locally Greedy. The convergence is faster when $\lambda$ is large because
        the activation becomes sparser and less coordinates need to be updated. Also, the locally greedy
        coordinate selection outperforms the greedy selection up to a certain point where the performances
        become similar as the sub-domain $\Sw$ becomes too small to be further partitioned with sub-domains
        of size $\Theta$.
    }
    \label{fig:annex:scaling_2d}
\end{figure}

\paragraph{Scaling of \mDICOD{} for 2D signals}
\autoref{fig:annex:scaling_2d} illustrates the scaling performance of \mDICOD{} on images. The average running times of \mDICOD{} are reported as a function of the number of workers $W$ for different regularization parameters $\lambda$ and for greedy and locally greedy selection. As for the 1D CSC resolution, the locally greedy selection consistently outperforms the greedy selection. This is particularly the case when using low number of workers $W$. Once the size of the sub-domains becomes smaller, the performances of both coordinate selection become closer as both algorithms become equivalent. Also, the convergence of \mDICOD{} is faster for large $\lambda$. This is expected, as in this case, the solution $Z^*$ is sparser and less coordinates need to be updated.

\subsection{Comparison with \citet{Skau2018}}

\begin{figure}[t]
    \centerline{\includegraphics[width=0.7\columnwidth]{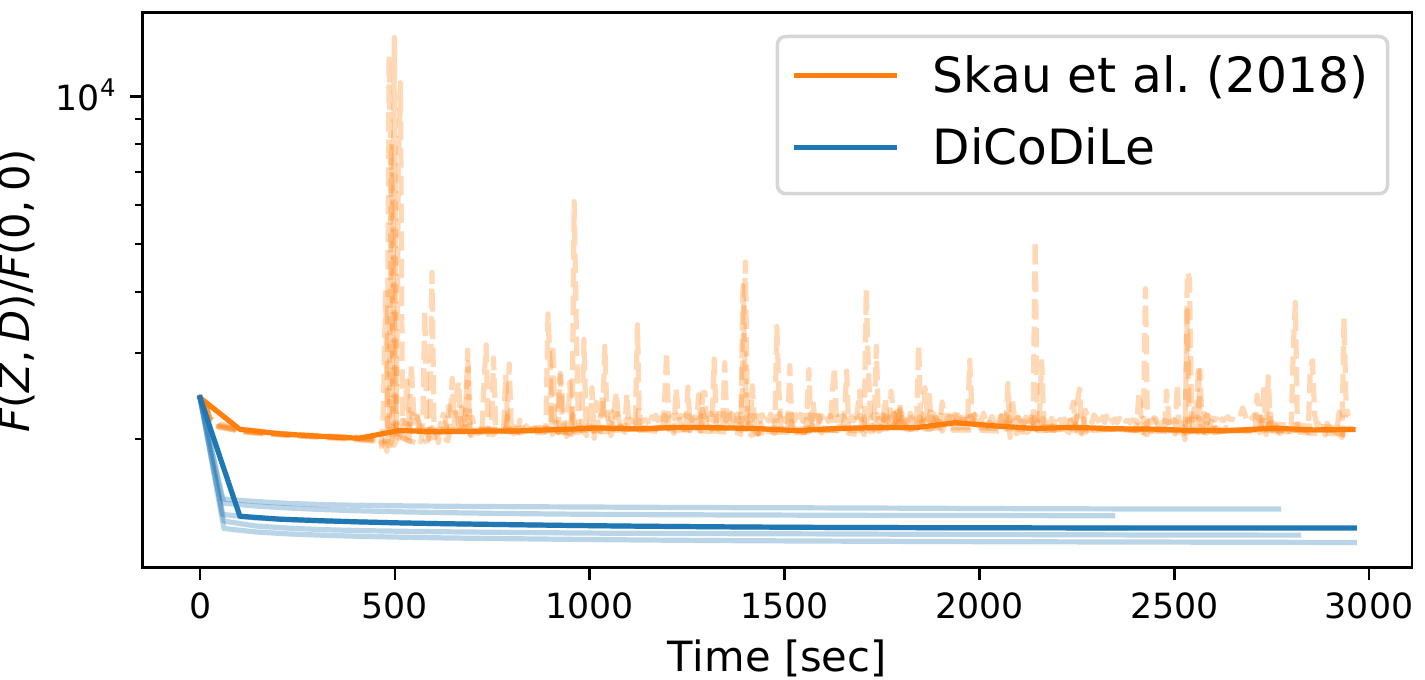}}
    \caption{
        Comparison of DiCoDiLe algorithm with Consensus ADMM dictionary learning proposed
        by \citet{Skau2018}. Algorithms were run 5 times with 36 workers on a random-patch
        of size $512\times 512$ taken from the Hubble Space Telescope GOODS South image.
        The solid lines denote the median curves obtained through interpolation.
    }
    \label{fig:annex:compare_cdl}
\end{figure}

\autoref{fig:annex:compare_cdl} displays the evolution of the objective function \autoref{eq:cdl} as a function of time for DiCoDiLe and the Consensus ADMM algorithm proposed by \citet{Skau2018}. Both algorithms were run on a single node with $36$ workers. We used the Hubble Space Telescope GOODS South image as input data. As the algorithm by \citet{Skau2018} raised a memory error with the full image, we used a random-patch of size $512\times 512$ as input data. Due to the non-convexity of the problem, the algorithms converge to different local minima, even when initialized with the same dictionary. The \autoref{fig:annex:compare_cdl} shows 5 runs of both algorithm, as well as the median curve, obtained by interpolation. We can see that DiCoDiLe outperforms the Consensus ADMM as it converges faster and to a better local minima. To ensure a fair comparison, the objective function is computed after projecting back the atoms of the dictionary $\pmb D_k$ to the $\ell_2$-ball by scaling it with $\|\pmb D_k\|_2^2$. This implies that we also needed to scale $Z$ by the inverse of the dictionary norm \ie $\frac{1}{\|\pmb d_k\|_2^2}$. This projection and scaling step is necessary for the solver as ADMM iterations do not guarantee that all iterates are feasible (\ie, satisfy the constraints).
This also explains the presence of several bumps in the evolution curve of the objective function.

\end{document}